\documentclass{article}

\PassOptionsToPackage{table,dvipsnames}{xcolor}

\usepackage[main, final]{neurips_2026}

\usepackage[utf8]{inputenc} 
\usepackage[T1]{fontenc}    
\usepackage{hyperref}       
\usepackage{url}            
\usepackage{booktabs}       
\usepackage{amsfonts}       
\usepackage{microtype}      
\usepackage{xcolor}         

\usepackage{graphicx}
\usepackage{fontawesome5}
\usepackage{subcaption}
\graphicspath{{./images/}}

\definecolor{rot}{RGB}{165,30,55} 

\usepackage{amsmath}
\usepackage{amssymb}
\usepackage{mathtools}
\usepackage{amsthm}

\usepackage{tcolorbox}
\usepackage{enumitem}
\usepackage{pifont}
\newcommand{\cmark}{\ding{51}}%
\newcommand{\xmark}{\ding{55}}%

\usepackage{array}
\usepackage{multirow}
\usepackage{longtable}
\usepackage{makecell}
\usepackage{colortbl}

\usepackage{adjustbox}
\usepackage{rotating}
\usepackage{wrapfig}
\usepackage{float}

\tcbuselibrary{skins,breakable}

\newtcolorbox{takeawaybox}{
  colback=white!90!gray,
  colframe=teal!60!black,
  arc=5pt,
  boxsep=5pt,
  left=5pt,
  right=10pt,
  top=2pt,
  bottom=3pt,
  boxrule=0.8pt,
  drop shadow=gray!50!white,
  enhanced jigsaw
}

\newcolumntype{C}[1]{>{\centering\arraybackslash}p{#1}}

\usepackage[capitalize,noabbrev]{cleveref}

\usepackage[textsize=tiny]{todonotes}

\theoremstyle{plain}

\theoremstyle{definition}

\theoremstyle{remark}

\title{Half-Truths Break Similarity-Based Retrieval}

\author{%
  Bora Kargi\thanks{Corresponding author: \texttt{kargibora@gmail.com}} \\
  University of T\"ubingen, T\"ubingen AI Center \\
  Konrad Zuse School of Excellence in Learning and Intelligent Systems (ELIZA) \\
  \And
  Arnas Uselis \\
  University of T\"ubingen, T\"ubingen AI Center \\
  \And
  Seong Joon Oh \\
  University of T\"ubingen, T\"ubingen AI Center \\
}

\begin{document}

\maketitle

\vspace{-12pt}
\begin{center}
  \hypersetup{hidelinks}
  \href{https://github.com/kargibora/CS-CLIP}{\faGithub\ Code}
  \qquad
  \href{https://huggingface.co/kbora/CS-CLIP}{%
    \raisebox{-0.5ex}{\includegraphics[height=2.5ex]{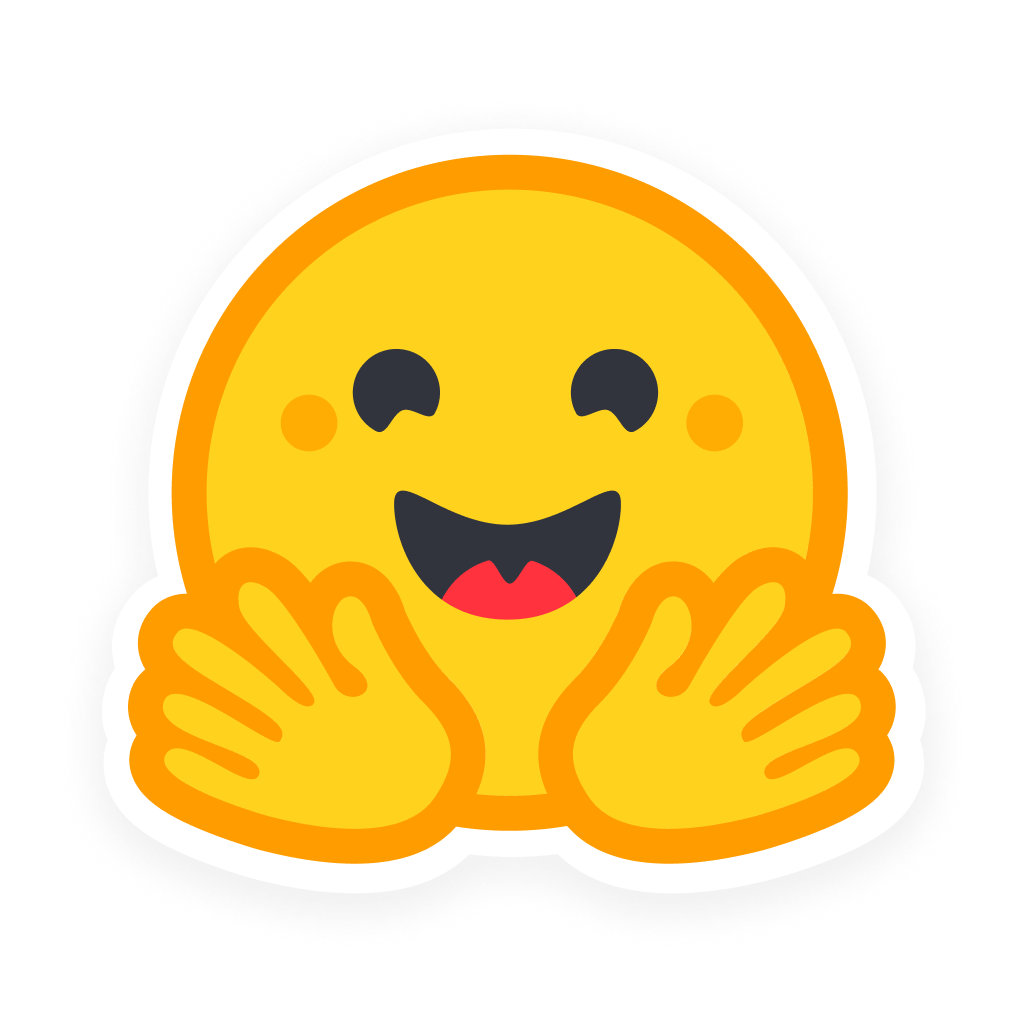}}%
    \ Checkpoints}
  \qquad
  \href{https://huggingface.co/datasets/kbora/Half-Truths}{%
    \raisebox{-0.5ex}{\includegraphics[height=2.5ex]{assets/resource-icons/huggingface.png}}%
    \ Dataset}
\end{center}
\vspace{12pt}

\begin{abstract}
When a text description is extended with an additional detail, image-text similarity should drop if that detail is wrong.
We show that CLIP-style dual encoders often violate this intuition: appending a plausible but incorrect object or relation to an otherwise correct description can increase the similarity score.
We call such cases \emph{half-truths}.
On COCO, CLIP prefers the correct shorter description only 34.0\% of the time, and performance drops to 27.8\% when the added detail is a relation.
We trace this vulnerability to weak supervision on caption parts: contrastive training aligns full sentences but does not explicitly enforce that individual entities and relations are grounded.
We propose CS-CLIP (Component-Supervised CLIP), which decomposes captions into entity and relation units, constructs a minimally edited foil for each unit, and fine-tunes the model to score the correct unit above its foil while preserving standard dual-encoder inference.
CS-CLIP raises half-truth accuracy to 76.4\% and improves average performance on established compositional benchmarks by 5.8 points, suggesting that reducing half-truth errors aligns with broader gains in compositional understanding.

\end{abstract}

\section{Introduction}

\begin{wrapfigure}{R}{0.55\textwidth}
  \centering
  \vspace{-2\baselineskip}
  \includegraphics[width=\linewidth]{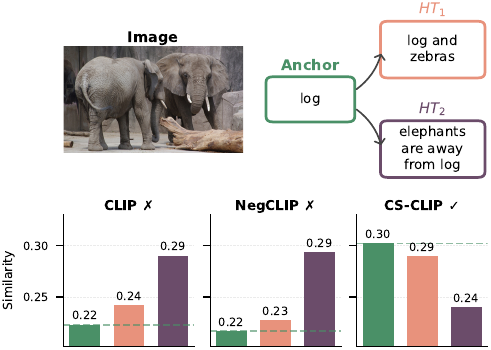}
  \caption{\textbf{The half-truth vulnerability.}
  Starting from a short, caption-supported description (the anchor), we form a \emph{half-truth}
  by adding one realistic but incorrect detail. CLIP and NegCLIP assign \emph{higher} similarity to the half-truth, while \textbf{CS-CLIP (ours)} correctly
  penalizes the incorrect addition.}
  \label{fig:halftruth_hero}
\end{wrapfigure}

When describing an image, adding a wrong detail should make the description \emph{less}
relevant, not more.
If ``a dog'' correctly describes an image, then ``a dog on a skateboard'' should score
\emph{lower} when the dog is not on a skateboard.
We show that CLIP-style dual encoders~\cite{radford2021learning} systematically violate this: appending a single
incorrect but plausible detail often \emph{increases} similarity.

We formalize this failure as the \textbf{half-truth vulnerability}.
Given an image, we start from a short, caption-supported \emph{anchor} description.
We then form a \emph{half-truth} by appending exactly one additional \emph{unit} that is
fluent and plausible in context but incorrect for the image.
The appended unit is either an \textbf{entity unit} (a wrong entity description) or a
\textbf{relation unit} (a wrong relation involving the anchor entity).
Figure~\ref{fig:halftruth_hero} illustrates both cases: an incorrect entity addition
(\emph{log} $\rightarrow$ \emph{log and zebras}) and an incorrect relation addition
(\emph{elephants away from the log} instead of near), where CLIP~\cite{radford2021learning}
and NegCLIP~\cite{yuksekgonul2022when} score the half-truth higher than the anchor,
while CS-CLIP correctly prefers the anchor.

This vulnerability is common.
On MS-COCO~\cite{lin2014coco}, CLIP ranks the anchor above the half-truth in only
$34.0\%$ of cases overall, and $27.8\%$ when the incorrect addition is a relation.
Even training with sentence-level hard negatives improves this only moderately:
NegCLIP achieves $53.8\%$ overall but remains below chance ($45.5\%$) for relation additions.
A single wrong detail, especially a relational one, is not reliably penalized.

This pattern resembles the conjunction fallacy, where adding a plausible detail
increases perceived likelihood even though conjunctions are more restrictive~\cite{tversky1983conjunction};
here the added detail is \emph{plausible but wrong}, yet similarity still rises. The key difference from prior work on compositional understanding is the operation:
existing benchmarks test whether models distinguish captions that \emph{swap or reorder}
information (e.g., ``red cat and blue dog'' vs.\ ``blue cat and red dog'')~\cite{yuksekgonul2022when,thrush2022winoground},
whereas half-truths test whether models \emph{penalize incorrect additions}.

\paragraph{Why does this happen?}
Contrastive training aligns images with \emph{full captions}.
This provides strong supervision at the sentence level, but weak supervision on the
individual \emph{units} that compose the caption's meaning.
As a result, similarity can be dominated by coarse overlap (e.g., detecting the right objects),
and an additional plausible unit can increase similarity even when it is incorrect, especially
for relations and role-sensitive structure that require verifying \emph{how} entities are composed.

\paragraph{Our approach: supervision on units.}
We propose \textbf{CS-CLIP} (\textbf{Component-Supervised CLIP}), which adds explicit
supervision on caption units during fine-tuning.
We parse each caption into \emph{entity units} (noun phrases with bound attributes)
and \emph{relation units} (directed relations between entities).
For each unit, we generate a minimally edited foil that preserves fluency and context
while changing meaning (e.g., ``brown horse'' $\to$ ``white horse'', or
``horse near barn'' $\to$ ``horse inside barn'').
During training, we contrast each correct unit against its foil, teaching the model
to respond to fine-grained compositional differences.
Critically, this supervision is applied \emph{only during training}, at test time,
CS-CLIP uses the same dual-encoder architecture and the same retrieval scoring as standard CLIP.

CS-CLIP achieves $76.4\%$ Half-Truth Accuracy on COCO with Qwen-generated foils,
compared to $34.0\%$ for CLIP and $53.8\%$ for NegCLIP.
It also reaches $67.2\%$ on COCO with Mistral-generated foils and $66.2\%$ on CC3M,
supporting transfer across foil generators and image domains.
It also improves on 16 established compositional benchmarks, achieving the best average
Image-to-Text accuracy ($57.8\%$) and Group Accuracy among evaluated models.

Our contributions are:
\begin{itemize}[leftmargin=*,noitemsep,topsep=1pt]
  \item \textbf{Diagnostic.}
  We introduce the half-truth diagnostic, which tests whether adding one incorrect
  detail increases similarity.
  CLIP fails this test in $66.0\%$ of COCO (Qwen) cases.

  \item \textbf{Method.}
  CS-CLIP fine-tunes CLIP with supervision on units: each entity and relation
  unit is contrasted against a matched foil.
  This achieves $76.4\%$ Half-Truth Accuracy (vs.\ $34.0\%$ for CLIP) while preserving
  standard dual-encoder inference.

  \item \textbf{Compositional benchmarks.}
  CS-CLIP achieves the best average I2T accuracy ($57.8\%$) and Group Accuracy across
  16 compositional benchmarks, outperforming CLIP by $5.8$ percentage points.
\end{itemize}

\section{Related Work}
\label{sec:related_work}

\textbf{Contrastive vision-language pretraining and compositional sensitivity.}
Contrastive dual encoders such as CLIP and ALIGN learn a shared embedding space for 
images and text, enabling efficient retrieval via a single similarity 
score~\cite{radford2021learning,jia2021scaling}.
A recurring concern is that this score can under-use linguistic structure: CLIP-style 
models behave like bag-of-words under minimal edits that change meaning, including 
attribute binding, word order, and relations~\cite{yuksekgonul2022when,thrush2022winoground}.
Recent analysis suggests this is largely a cross-modal issue: binding cues may be 
present within modalities but not reliably reflected in cosine 
alignment~\cite{koishigarina2026clipbow}; other work reports scoring biases that 
favor central objects or frequent attribute-object pairings~\cite{schrodi2025two,tang2023}.
To evaluate this, controlled-edit benchmarks test whether models distinguish correct 
image-text pairs from minimally edited alternatives, including ARO, Winoground, 
SugarCrepe, and VALSE~\cite{yuksekgonul2022when,thrush2022winoground,hsieh2023sugarcrepe,parcalabescu2022valse}, 
as well as targeted tests for specific 
phenomena~\cite{burapacheep2024colorswap,kamath2023whats,paiss2023countbench}.

\textbf{Adding information and our diagnostic.}
Beyond swaps and reorderings, several works study how model confidence changes as more 
information is added to an input.
HiMo-CLIP targets monotonic alignment under correct 
additions~\cite{wu2026himoclip}, while confidence inflation 
patterns in LLM reasoning show that added plausible context can inflate confidence 
under misleading inputs~\cite{suri2024decision,richardson2025fallacypatterns,jiang-etal-2024-peek}.
Our work is complementary: we test whether similarity properly penalizes incorrect 
additions through the half-truth diagnostic, which appends one realistic but incorrect 
detail to a correct description (Section~\ref{sec:motivation:half_truth}).

\textbf{Training methods for compositional sensitivity.}
Many methods strengthen sensitivity through data-level signals: sentence-level hard 
negatives that edit objects, attributes, or 
relations~\cite{yuksekgonul2022when,zhang2024contrasting,peleg2025clic,singh2025conclip,patel2024tripletclip}, 
scene-graph perturbations~\cite{singh2023mosaic}, or caption 
synthesis~\cite{fan2023improving,doveh2023dac,castro2024clove,doveh2023svlc,nayak2023learning}.
Other approaches add architectural components or auxiliary objectives such as region 
alignment~\cite{zhong2022regionclip,huang2024structureclip}, auxiliary 
modules~\cite{yao2021filip,goel2022cyclip,abdollahi2024comalign,oh2024preserving,kwon2025readclip}, 
or multi-stage training~\cite{hu2025decoupledgloballocalalignmentimproving}, while 
some modify inference through decomposition~\cite{jiang2024comclip,miranda2025adding,menon2023visual}.
We build on sentence-level negatives but add targeted unit-level supervision: for each 
caption, we parse entity and relation units and contrast each against a minimally 
edited foil.
This provides direct compositional pressure on individual caption parts while 
maintaining the standard dual-encoder architecture (Section~\ref{sec:method}).
\section{The Half-Truth Vulnerability}
\label{sec:motivation:half_truth}

\subsection{Motivation}

Prior work on compositional sensitivity in CLIP-style models has mainly studied
minimal edits that \emph{rearrange} existing information, such as swapping
attributes or changing role assignments~\cite{yuksekgonul2022when,thrush2022winoground}.
We study a complementary retrieval failure: \emph{adding} incorrect information.

In retrieval, users often start from a short but correct description and then
refine it by appending more details.
If the added detail is wrong, the refined query should become \emph{less} relevant
to the image, not more.
Yet CLIP-style similarity can violate this basic monotonicity intuition:
adding one plausible but false detail can increase the score.
We call this failure the \textbf{half-truth vulnerability}.

\subsection{Half-Truth Diagnostic: Metric and Construction}

\begin{wrapfigure}{R}{0.50\textwidth}
  \centering
  \vspace{-0.65\baselineskip}
  \includegraphics[width=\linewidth]{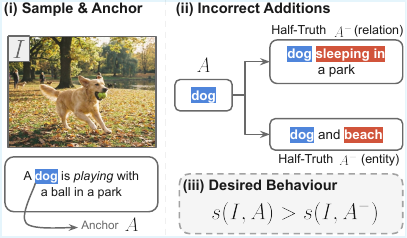}
  \caption{\textbf{Half-truth construction.}
  Parse captions into units and generate foils via minimal edits, then sample an anchor $A$ and append one foil to form half-truth $A^{-}$.}
  \label{fig:halftruth_construction}
\end{wrapfigure}

\paragraph{Anchor vs.\ half-truth.}
Given an image $I$, we compare a short correct description, the \textbf{anchor}
$A$, with a \textbf{half-truth} $A^{-}$ formed by appending exactly one additional
detail that is plausible in context but incorrect for the image.
The desired ordering is simple: an incorrect refinement should not outrank a
correct but incomplete one.
We therefore report \textbf{Half-Truth Accuracy} as the fraction of cases where
the model prefers the anchor over the half-truth:
\begin{equation}
\text{Acc}_{\mathrm{HT}} \;=\; \Pr\big[s(I, A) > s(I, A^{-})\big],
\label{eq:anchor_over_incorrect_ext}
\end{equation}
where $s(\cdot,\cdot)$ is the image--text similarity score (higher is better).
Random choice corresponds to $50\%$.
Values below $50\%$ mean that the model often assigns higher similarity to the
incorrect refinement than to the correct anchor.

\paragraph{Diagnostic construction.}
We construct COCO comparisons under the Karpathy partition~\cite{karpathy2015deep}
using Qwen3-8B and Mistral-8B foil generators, and CC3M comparisons using Qwen3-8B.
We first construct candidate half-truths automatically from captions, then use
manual verification to retain only clean examples for the main paper results.
Figure~\ref{fig:halftruth_construction} illustrates the construction.

  We parse each caption $T$ into two kinds of textual units using a text-only LLM pipeline (Appendix~\ref{app:half_truth_construction}): \textbf{entity units} $
  \mathcal{E}(T)$, noun phrases including attributes and quantifiers (e.g., ``brown horse'', ``three dogs''), and \textbf{relation units} $\mathcal{R}(T)$, directed
  relations between two entities (e.g., ``person riding horse'', ``ball in park'').

For each unit, the same pipeline proposes minimally edited \emph{matched foils}, incorrect
variants that are plausible in context. For entity units, foils change either the object (e.g., ``brown horse'' $\rightarrow$
``brown giraffe'') or an attribute (e.g., ``brown horse'' $\rightarrow$ ``white horse'').

For relation units, foils change the predicate, swap arguments, or replace one entity.
The anchor $A$ is sampled from the entity units $\mathcal{E}(T)$, and a half-truth
$A^{-}$ is formed by appending exactly one foil from either $\mathcal{E}(T)$ or
$\mathcal{R}(T)$.
When appending a relation unit, the relation introduces a second entity and describes
an interaction or spatial relationship between it and the anchor entity.
This construction ensures that half-truths differ from anchors by a single, targeted error.

\paragraph{Verification.}
We screen candidate foils with Qwen3-VL-30B and use Claude Opus 5 to review them
and propose alternatives within the same foil category. The authors manually check
the final evaluation candidates, including replacements, against their images
and discard foils judged correct or ambiguous. The evaluation contains
1,437 comparisons: 757 entity and 680 relation additions across COCO (Qwen),
COCO (Mistral), and CC3M (Qwen). Manual verification applies only to evaluation;
training foils are generated and filtered automatically.
Appendix~\ref{app:half_truth_per_condition} gives the source-wise counts and training-foil audit.

\subsection{Half-Truth Vulnerability on COCO}
\begin{wrapfigure}{R}{0.60\textwidth}
  \centering
  \includegraphics[width=\linewidth]{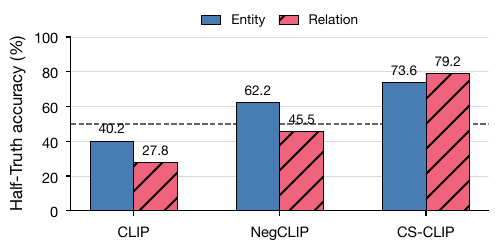}
  \caption{\textbf{Half-Truth Accuracy on COCO (Qwen).} Entity versus relation additions; dashed line indicates chance.}
  \label{fig:motivation_bars}
\end{wrapfigure}
On COCO (Qwen), CLIP reaches 34.0\% overall accuracy: 40.2\% for entity additions
and 27.8\% for relation additions (Figure~\ref{fig:motivation_bars}).
NegCLIP improves these to 62.2\% and 45.5\%, respectively, leaving a substantial
relation gap. CS-CLIP reaches 73.6\% on entities and 79.2\% on relations.

\paragraph{Why unit-level supervision?}
Half-truths test whether similarity responds to the specific added detail.
This motivates contrasting each caption unit against a matched foil during
training, alongside the full-caption objective. Section~\ref{sec:exp:ablations}
compares sentence-level and unit-level use of these foils.

\newcommand{\kappaI}{\kappa}

\section{Method: CS-CLIP}
\label{sec:method}

The half-truth diagnostic reveals that models struggle when incorrect information is 
added to correct descriptions.
To address this, we introduce unit-level supervision during fine-tuning.
Rather than training on anchor-vs-half-truth comparisons directly, we provide supervision 
at the unit level: for each caption unit, we contrast the correct unit against a minimally 
edited foil (e.g., ``brown horse'' vs.\ ``white horse'').
This targets the underlying weakness, sensitivity to compositional structure, making the 
approach general beyond the half-truth setting.

Standard CLIP training aligns each image with its full caption using a global contrastive 
objective, which can leave the similarity score insensitive to individual caption parts 
(Section~\ref{sec:motivation:half_truth}). 
To address this, we propose CS-CLIP (Component-Supervised CLIP), which adds targeted 
unit-level supervision during fine-tuning while keeping the dual-encoder architecture 
and standard cosine scoring unchanged at test time.

\begin{figure*}[t]
  \centering
  \includegraphics[width=0.95\textwidth]{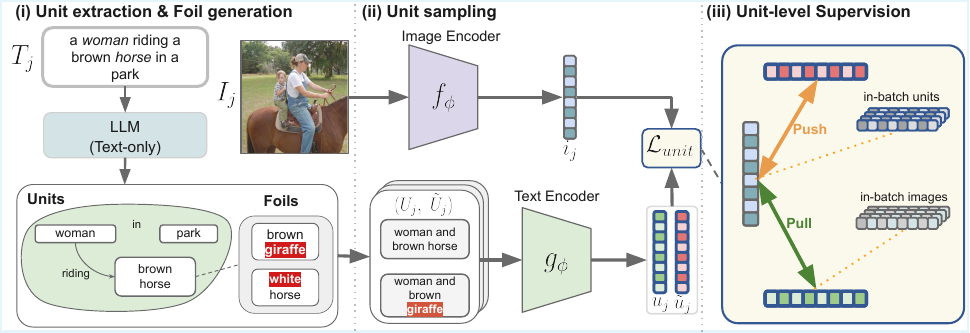}
    \caption{\textbf{CS-CLIP training pipeline.}
  (i) A text-only LLM extracts entity and relation units from each caption and creates matched foils via minimal edits.
  (ii) For each image-caption pair, we sample unit--foil pairs and encode the image and unit texts with the dual encoder.
  (iii) The unit loss scores the correct unit above its matched foil and other in-batch units, while the standard sentence-level contrastive loss is
  applied in parallel.}
  \label{fig:pipeline}
\end{figure*}

For each image-caption pair, we parse the caption into entity units and relation units 
(defined in Section~\ref{sec:motivation:half_truth}), sample a unit with its matched 
foil, and train the image embedding to score the correct unit higher than the foil. 
This direct supervision on caption parts makes similarity more sensitive to individual 
details while preserving full-caption alignment through a parallel sentence-level loss.
Figure~\ref{fig:pipeline} illustrates the complete training pipeline.

\subsection{Unit Sampling and Encoding}
\label{sec:method:units}

We reuse the text-only parsing pipeline from Section~\ref{sec:motivation:half_truth} 
to extract entity units $\mathcal{E}(T)$ and relation units $\mathcal{R}(T)$ from each 
caption $T$, along with matched foils for each unit (Appendix~\ref{app:llm_parsing} 
provides implementation details).

\textbf{Sampling strategy.}
For each image-caption pair, we sample one unit $U$ from either entities or relations.
With probability $p_{\mathrm{rel}}$, we sample a relation unit $U \in \mathcal{R}(T)$; 
otherwise, we sample an entity unit $U \in \mathcal{E}(T)$, optionally forming 
conjunctions of up to $K{=}2$ entities.
We sample the unit's matched foil $\tilde{U}$ and encode both with the text encoder 
$g_\phi(\cdot)$ as noun phrases (entities) or predicate phrases (relations), yielding 
normalized embeddings.
We investigate other sampling strategies in Appendix~\ref{app:ablations_full}.

\subsection{Training Objective}
\label{sec:method:objective}

\textbf{Sentence-level loss ($\mathcal{L}_{\mathrm{global}}$).}
We use a standard global image-caption contrastive objective with NegCLIP-style hard 
negatives~\cite{yuksekgonul2022when} (Appendix~\ref{app:global_loss}), which we combine 
with unit-level supervision to improve sensitivity to compositional structure.

\textbf{Unit-level loss ($\mathcal{L}_{\mathrm{unit}}$).}
For unit supervision, we use the same cosine scoring as CLIP with a learnable temperature $\tau$.
Since embeddings are normalized, cosine similarity equals a dot product, and we define $\kappa(\mathbf{x},\mathbf{y})=\exp(\mathbf{x}^\top \mathbf{y}/\tau)$.

For each image $I_i$, we sample $N_u$ unit/foil pairs $\{(U_{i,k},\tilde{U}_{i,k})\}_{k=1}^{N_u}$.
Let $\mathbf{v}_i$ denote the normalized image embedding, and let $\mathbf{u}_{i,k}$
and $\tilde{\mathbf{u}}_{i,k}$ denote the normalized embeddings of the $k$-th unit and
its matched foil for image $i$.
For a fixed unit index $k$, the image-to-unit and symmetric unit-to-image losses are:
\begin{equation}
\mathcal{L}^{(k)}_{I\rightarrow U}
= -\frac{1}{B}\sum_{i=1}^{B}\log
\frac{\kappa(\mathbf{v}_i,\mathbf{u}_{i,k})}
{\sum_{j=1}^{B}\kappa(\mathbf{v}_i,\mathbf{u}_{j,k})+\kappa(\mathbf{v}_i,\tilde{\mathbf{u}}_{i,k})},
\;
\mathcal{L}^{(k)}_{U\rightarrow I}
= -\frac{1}{B}\sum_{i=1}^{B}\log
\frac{\kappa(\mathbf{v}_i,\mathbf{u}_{i,k})}
{\sum_{j=1}^{B}\kappa(\mathbf{v}_j,\mathbf{u}_{i,k})}.
\label{eq:unit_losses_main}
\end{equation}
$\mathcal{L}^{(k)}_{I\rightarrow U}$ trains the image to score the correct unit $U_{i,k}$ higher than (i) its matched foil $\tilde{U}_{i,k}$ and (ii) in-batch units from other images.
We exclude foils from $\mathcal{L}^{(k)}_{U\rightarrow I}$ since each foil is tied to a specific image-unit pair; the symmetric term provides bidirectional regularization. We average over the $N_u$ sampled unit indices: $\mathcal{L}_{\mathrm{unit}}=\tfrac{1}{2N_u}\sum_{k=1}^{N_u}(\mathcal{L}^{(k)}_{I\rightarrow U}+\mathcal{L}^{(k)}_{U\rightarrow I})$.

\textbf{Final objective.}
We combine global alignment with unit-level supervision:
\begin{equation}
\mathcal{L}_{\mathrm{CS}}
= \mathcal{L}_{\mathrm{global}} + \lambda_u \, \mathcal{L}_{\mathrm{unit}}.
\label{eq:csclip_loss_main}
\end{equation}
This objective keeps standard CLIP inference unchanged, but makes the learned similarity 
more sensitive to caption structure by training the model to distinguish correct caption 
units from their matched foils.

\section{Experiments}

\begin{table*}[t]
\centering
\scriptsize
\setlength{\tabcolsep}{3pt}
\renewcommand{\arraystretch}{1.08}
\caption{\textbf{Half-Truth Accuracy across foil generators and image domains.}
We report percentages on 509 COCO (Qwen), 454 COCO (Mistral), and 474 CC3M (Qwen) manually verified comparisons. All is the sample-weighted entity/relation average. FSC-CLIP is trained on CC3M; the other fine-tuned models use COCO. Higher is better.}
\label{tab:halftruth_overall_and_split}
\begin{tabular*}{\textwidth}{@{\extracolsep{\fill}}l|ccc|ccc|ccc@{}}
\toprule
& \multicolumn{3}{c|}{\textbf{COCO (Qwen)}} & \multicolumn{3}{c|}{\textbf{COCO (Mistral)}} & \multicolumn{3}{c}{\textbf{CC3M (Qwen)}} \\
\textbf{Model} & All & Entity & Relation & All & Entity & Relation & All & Entity & Relation \\
\midrule
CLIP & 34.0 & 40.2 & 27.8 & 28.6 & 34.5 & 21.3 & 36.1 & 46.6 & 24.2 \\
NegCLIP & 53.8 & 62.2 & 45.5 & 45.2 & 54.8 & 33.2 & 55.3 & 67.7 & 41.3 \\
DeGLA & 46.2 & 59.1 & 33.3 & 40.1 & 53.2 & 23.8 & 52.3 & 64.9 & 38.1 \\
ReadCLIP & 58.0 & 66.1 & 49.8 & 50.9 & 60.3 & 39.1 & 60.1 & 65.7 & 53.8 \\
FSC-CLIP (CC3M) & 61.3 & 70.9 & 51.8 & 54.0 & 58.3 & 48.5 & \textbf{67.5} & \textbf{76.5} & 57.4 \\
\rowcolor{gray!10}CS-CLIP & \textbf{76.4} & \textbf{73.6} & \textbf{79.2} & \textbf{67.2} & \textbf{70.2} & \textbf{63.4} & 66.2 & 68.9 & \textbf{63.2} \\
\bottomrule
\end{tabular*}
\end{table*}

We evaluate CS-CLIP in three settings.
First, we measure robustness to incorrect additions using the half-truth diagnostic (Section~\ref{sec:motivation:half_truth}).
Second, we evaluate compositional generalization on established controlled-edit benchmarks.
Third, we report downstream zero-shot classification and image--text retrieval results, and we include ablations to isolate the effect of each design choice.

\textbf{Implementation and training.}
We fine-tune CLIP using the OpenCLIP codebase~\cite{cherti2023openclip} with OpenAI CLIP initialization.
Unless stated otherwise, we use ViT-B/32~\cite{radford2021learning}. We fine-tune on MS-COCO under the Karpathy split~\cite{karpathy2015deep}
for 25 epochs with batch size 128 on 8$\times$A100 GPUs, using AdamW (lr $5\times10^{-6}$, weight decay $10^{-2}$).
We use cosine warm-up for the first three epochs followed by cosine decay.
For the unit-level objective, we sample $N_u=2$ unit/foil pairs per image per step and use a learnable temperature $\tau$ initialized from the pretrained CLIP temperature.
We report results from the final training checkpoint.

\textbf{Training objective.}
We set the unit-loss weight to $\lambda_u=0.5$. For global caption-level negatives we follow NegCLIP-style training~\cite{yuksekgonul2022when},
using in-batch captions plus shuffled synthetic hard negatives. 
Caption parsing and unit sampling follow Section~\ref{sec:method:units}; implementation details are in Appendix~\ref{app:half_truth_construction}.

\subsection{Half-Truth Vulnerability}
\label{sec:exp:halftruth}
Table~\ref{tab:halftruth_overall_and_split} reports Half-Truth Accuracy on the three
evaluation sources. COCO (Mistral) changes the foil generator, while CC3M (Qwen)
changes the image domain.

\textbf{COCO results.}
CS-CLIP reaches 76.4\% on COCO (Qwen), improving over CLIP by 42.4 points
and FSC-CLIP by 15.1 points. The largest gain is on relations: CLIP reaches
27.8\%, NegCLIP 45.5\%, and FSC-CLIP 51.8\%, compared to 79.2\% for CS-CLIP.
CS-CLIP also leads on entity additions at 73.6\%.

\textbf{Cross-generator and cross-domain transfer.}
With Mistral-generated COCO foils, CS-CLIP reaches 67.2\%, compared to
28.6\% for CLIP and 54.0\% for FSC-CLIP. On CC3M, CS-CLIP improves over
CLIP from 36.1\% to 66.2\%; CC3M-trained FSC-CLIP leads overall at 67.5\%.
CS-CLIP remains strongest on relations among these baselines across all three
sources (79.2\%, 63.4\%, and 63.2\%). Its mean accuracy across sources is
69.9\%, versus 60.9\% for FSC-CLIP.

\textbf{Correct refinements.}
CS-CLIP also ranks a correct refinement above the anchor and the half-truth below
it more often than the displayed baselines: Correct Ordering Rate is 58.2\%,
50.4\%, and 42.2\% on COCO (Qwen), COCO (Mistral), and CC3M (Qwen), respectively
(Appendix~\ref{app:half_truth_ordering}). This supports sensitivity to the
added content rather than a uniform preference for shorter descriptions.

\subsection{Compositional Understanding}
\label{sec:exp:compositionality}

\begin{wrapfigure}{R}{0.5\textwidth}
    \centering
    \vspace{-2\baselineskip}
    \includegraphics[width=\linewidth]{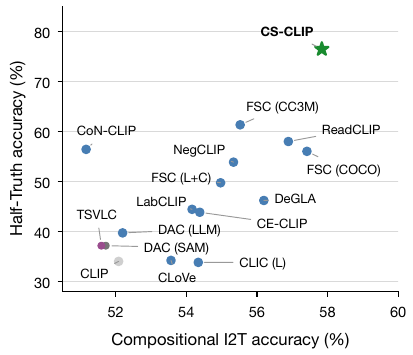}
    \caption{\textbf{Compositional I2T vs.\ Half-Truth Accuracy.}
    Sixteen models; Half-Truth Accuracy is measured on COCO (Qwen). L: LAION; C: COCO.}
    \label{fig:comp_vs_halftruth_scatter}
  \end{wrapfigure}
  
  The half-truth diagnostic isolates sensitivity to a single incorrect added
  detail. We next test whether CS-CLIP improves broader compositional
  understanding on controlled-foil benchmarks covering attribute binding,
  relations, and linguistic variation, including ARO~\cite{yuksekgonul2022when},
  SugarCrepe~\cite{hsieh2023sugarcrepe}, and
  Winoground~\cite{thrush2022winoground}. We report I2T, T2I, and, for paired
  datasets, Group Accuracy (Grp); full protocols are in
  Appendix~\ref{app:compositional:metrics}.



\textbf{Overall compositional performance.}
Figure~\ref{fig:comp_vs_halftruth_scatter} plots average compositional I2T accuracy against Half-Truth Accuracy across models.
CS-CLIP achieves the best performance on both metrics: \textbf{57.8\%} compositional I2T ($+$5.8~pp over CLIP; Table~\ref{tab:summary_compositional}) and \textbf{76.4\%} $\mathrm{Acc}_{\mathrm{HT}}$ on COCO (Qwen).
The gains on independently constructed benchmarks support transfer beyond the half-truth diagnostic.

\begin{wrapfigure}{L}{0.5\textwidth}
    \centering
    \vspace{-0.35\baselineskip}
    \includegraphics[width=\linewidth]{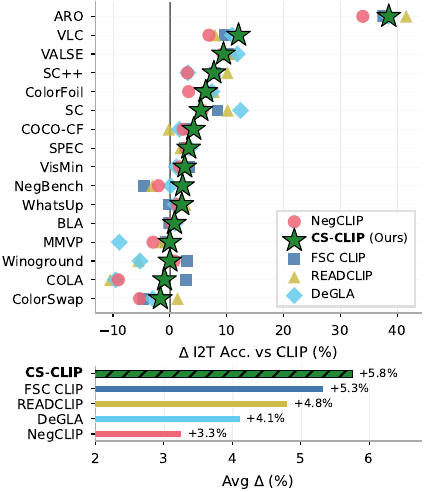}
    \caption{\textbf{Per-benchmark compositional gains.}
    $\Delta$I2T relative to zero-shot CLIP by benchmark and on average.}
    \label{fig:comp_dataset_delta}
  \end{wrapfigure}

\textbf{Dataset-wise breakdown.}
CS-CLIP improves over NegCLIP on 14/16 benchmarks (Figure~\ref{fig:comp_dataset_delta}) 
and outperforms the next best COCO-trained baselines by $+$0.4~pp (FSC-CLIP), $+$0.9~pp (ReadCLIP), and $+$1.6~pp (DeGLA). 
CS-CLIP preserves the dual-encoder architecture and test-time scoring. 
CS-CLIP achieves the best VL Checklist \cite{zhao2022vlchecklist} score (79.2\%), 
a systematic audit of objects, attributes, and relations.

 \textbf{Paired benchmarks.}
  On paired compositional datasets, CS-CLIP also improves Text-to-Image accuracy
  and achieves the best average Group Accuracy among evaluated models, indicating
  that gains do not come at the expense of weaker matching in the opposite
  retrieval direction. Full bidirectional results are in
  Appendix~\ref{app:compositional:bidirectional}.

\textbf{Main takeaway.}
CS-CLIP achieves the best average compositional I2T accuracy (57.8\%, $+$5.8~pp over CLIP) and the best average Group Accuracy among evaluated models.
Improvements are consistent across datasets and aligned with half-truth robustness, demonstrating that unit-level supervision addresses compositional understanding broadly rather than overfitting to specific tests.

\subsection{Downstream Performance}
\label{sec:exp:downstream}

We evaluate downstream performance with CLIPBench~\cite{cherti2023clipbenchmark} on zero-shot classification (ImageNet variants and standard recognition datasets) and image-text retrieval (COCO and Flickr8k~\cite{plummer2015flickr30k}).
Full per-dataset results and evaluation details are in Appendix~\ref{app:downstream}.

\textbf{Zero-shot classification.}
Table~\ref{tab:downstream_zs_combined} shows zero-shot classification results.
Compared to CLIP, CS-CLIP changes average accuracy modestly (Acc@1: \textbf{63.6} $\rightarrow$ \textbf{59.9}; Acc@5: \textbf{86.5} $\rightarrow$ \textbf{84.6}).
This drop is comparable to other COCO-fine-tuned baselines (NegCLIP: 58.2\%, FSC-CLIP: 61.2\%), indicating that unit-level supervision trades off zero-shot accuracy similarly to other fine-tuning approaches.
The unit-only ablation reaches 61.8\%, close to plain COCO fine-tuning at 62.0\%;
global hard-negative training produces a larger drop (NegCLIP: 58.2\%).

\textbf{Retrieval tasks.}
Table~\ref{tab:downstream_retrieval} reports Recall@1 on COCO and Flickr8k \cite{plummer2015flickr30k}.
Since CS-CLIP is fine-tuned on COCO, we treat retrieval as an internal comparison among COCO-fine-tuned models.
CS-CLIP achieves the \textbf{best average T2I Recall@1} (\textbf{71.7}, $+$5.8~pp over NegCLIP and $+$1.9~pp over FSC-CLIP) and ties for the \textbf{best average I2T} (\textbf{56.8}), demonstrating that unit-level supervision improves retrieval alignment while maintaining competitive performance in both directions.

\subsection{Ablations}
\label{sec:exp:ablations}
\begin{table}[t]
\centering
\caption{\textbf{Backbone scaling and fine-tuning strategy.}
For ViT-B/32, we compare \textbf{Full FT} (updating both encoders) against \textbf{Text-only FT} and \textbf{Image-only FT}.
We also report backbone scaling under Full FT.}
\label{tab:abl_backbone_and_ftstrategy}

\scriptsize
\setlength{\tabcolsep}{2.1pt}
\renewcommand{\arraystretch}{1.06}

\begin{adjustbox}{max width=\columnwidth}
\begin{tabular*}{\columnwidth}{@{\extracolsep{\fill}}l|ccc|ccc|c|cc@{}}
\toprule
\textbf{Model} &
\multicolumn{3}{c|}{\textbf{Compositionality}} &
\multicolumn{3}{c|}{\textbf{Half-Truth}} &
\multicolumn{1}{c|}{\textbf{ZS}} &
\multicolumn{2}{c}{\textbf{Retr}} \\
\cmidrule(r){2-4}\cmidrule(lr){5-7}\cmidrule(lr){8-8}\cmidrule(l){9-10}
& \textbf{I2T} & \textbf{T2I} & \textbf{Grp} & \textbf{All} & \textbf{Ent} & \textbf{Rel} & \textbf{Avg} & \textbf{T2I} & \textbf{I2T} \\
\midrule

\multicolumn{10}{@{}l}{\textit{ViT-B/32 (fine-tuning strategy)}} \\
\rowcolor{gray!10}\textbf{ViT-B/32 Full FT (Ours)} & 57.8 & 39.8 & 27.8 & 76.4 & 73.6 & 79.2 & 59.9 & \textbf{71.7} & \textbf{56.8} \\
ViT-B/32 Text-only FT & 55.9 & 38.0 & 26.0 & 62.5 & 74.4 & 50.6 & \textbf{61.4} & 69.5 & 53.0 \\
ViT-B/32 Image-only FT & 56.2 & 37.9 & 27.4 & 59.5 & 72.0 & 47.1 & 58.6 & 65.6 & 54.0 \\
\midrule

\multicolumn{10}{@{}l}{\textit{Backbone scaling (Full FT)}} \\
ViT-B/16 & 58.5 & 39.8 & 28.7 & 78.8 & 85.0 & 72.5 & 62.5 & 74.8 & 61.6 \\
ViT-L/14 & \textbf{59.5} & \textbf{42.4} & \textbf{31.4} & 80.4 & 86.6 & 74.1 & \textbf{65.9} & \textbf{79.9} & \textbf{67.1} \\
\bottomrule
\end{tabular*}
\end{adjustbox}
\end{table}

\begin{table}[t]
\centering
\caption{\textbf{Effect of the unit-loss weight $\lambda_u$.}
We vary $\lambda_u$ while keeping all other settings fixed.}
\label{tab:abl_lambda_singlecol}

\scriptsize
\setlength{\tabcolsep}{2.1pt}
\renewcommand{\arraystretch}{1.06}

\begin{adjustbox}{max width=\columnwidth}
\begin{tabular*}{\columnwidth}{@{\extracolsep{\fill}}l|ccc|ccc|c|cc@{}}
\toprule
\textbf{$\lambda_u$} &
\multicolumn{3}{c|}{\textbf{Compositionality}} &
\multicolumn{3}{c|}{\textbf{Half-Truth}} &
\multicolumn{1}{c|}{\textbf{ZS}} &
\multicolumn{2}{c}{\textbf{Retr}} \\
\cmidrule(r){2-4}\cmidrule(lr){5-7}\cmidrule(lr){8-8}\cmidrule(l){9-10}
& \textbf{I2T} & \textbf{T2I} & \textbf{Grp} & \textbf{All} & \textbf{Ent} & \textbf{Rel} & \textbf{Avg} & \textbf{T2I} & \textbf{I2T} \\
\midrule
0.10 & 56.9 & 38.4 & 26.4 & 73.1 & 70.9 & 75.3 & \textbf{60.7} & 70.8 & 56.7 \\
0.25 & 56.7 & 39.4 & 27.2 & 75.4 & 73.6 & 77.3 & 59.3 & 71.0 & 56.5 \\
\rowcolor{gray!10}\textbf{0.50 (Ours)} & \textbf{57.8} & \textbf{39.8} & 27.8 & 76.4 & 73.6 & 79.2 & 59.9 & \textbf{71.7} & 56.8 \\
0.75 & 57.7 & 39.3 & \textbf{27.9} & 77.6 & 73.6 & 81.6 & 59.8 & 71.1 & \textbf{56.9} \\
1.00 & 57.2 & 38.5 & 27.2 & 78.0 & 76.0 & 80.0 & 60.6 & 71.1 & 56.4 \\
\bottomrule
\end{tabular*}
\end{adjustbox}
\end{table}

\begin{table}[t]
\centering
\scriptsize
\setlength{\tabcolsep}{2pt}
\caption{\textbf{Training-signal ablations.} G: global sentence negatives; U: unit supervision; F: matched unit foils. Half-Truth accuracy uses COCO (Qwen); ZS is average zero-shot classification accuracy.}
\label{tab:abl_loss_design_checks}
\begin{tabular*}{\textwidth}{@{\extracolsep{\fill}}cl|ccc|ccc|ccc@{}}
\toprule
& & \multicolumn{3}{c|}{Compositionality} & \multicolumn{3}{c|}{Half-Truth} & \multicolumn{3}{c}{Downstream} \\
Variant & G/U/F & I2T & T2I & Group & All & Entity & Relation & ZS & Retr. T2I & I2T \\
\midrule
1 & \xmark\,\xmark\,\xmark & 52.8 & 39.4 & 27.2 & 43.8 & 62.6 & 25.1 & 62.0 & 71.7 & 56.7 \\
2 & \cmark\,\xmark\,\xmark & 55.3 & 38.5 & 25.7 & 53.8 & 62.2 & 45.5 & 58.2 & 65.9 & 52.7 \\
3 & \xmark\,\cmark\,\xmark & 52.5 & 39.8 & 28.0 & 46.0 & 66.9 & 25.1 & 61.8 & 71.3 & 56.2 \\
4 & \xmark\,\cmark\,\cmark & 54.8 & 39.1 & 28.3 & 76.2 & 74.8 & 77.6 & 60.7 & 70.7 & 56.7 \\
5 & \cmark\,\cmark\,\xmark & 55.2 & 38.6 & 27.3 & 47.7 & 65.0 & 30.6 & 59.6 & 70.9 & 56.6 \\
\rowcolor{gray!10}6 & \cmark\,\cmark\,\cmark & 57.8 & 39.8 & 27.8 & 76.4 & 73.6 & 79.2 & 59.9 & 71.7 & 56.8 \\
\bottomrule
\end{tabular*}
\end{table}

\textbf{Backbone and fine-tuning strategy.}
Table~\ref{tab:abl_backbone_and_ftstrategy} shows that updating both encoders
reaches 79.2\% relation accuracy, compared to 50.6\% for text-only and 47.1\%
for image-only tuning. Larger backbones improve overall Half-Truth accuracy
and downstream performance.

\textbf{Unit-loss weight.}
Increasing $\lambda_u$ from 0.10 to 0.75 raises relation accuracy from 75.3\%
to 81.6\%, while compositional I2T varies less (Table~\ref{tab:abl_lambda_singlecol}).
We use $\lambda_u=0.5$ to balance these metrics.

\textbf{Training signals.}
Table~\ref{tab:abl_loss_design_checks} separates global negatives, unit supervision,
and matched unit foils. With global negatives and unit supervision, adding
matched foils raises Half-Truth accuracy from 47.7\% to 76.4\%, and relation
accuracy from 30.6\% to 79.2\%. Unit supervision alone yields 61.8\% zero-shot accuracy,
close to plain COCO fine-tuning at 62.0\%. Combining all signals gives 57.8\%
compositional I2T and 59.9\% zero-shot accuracy.

\begin{table}[t]
\centering
\small
\caption{\textbf{Foil placement and generator controls.} Half-Truth columns report entity/relation accuracy (\%); Comp. is average compositional I2T accuracy.}
\label{tab:foil_controls}
\setlength{\tabcolsep}{3pt}
\begin{tabular*}{\textwidth}{@{\extracolsep{\fill}}lrrrr@{}}
\toprule
Model & COCO (Qwen) & COCO (Mistral) & CC3M (Qwen) & Comp. \\
\midrule
NegCLIP & 62.2 / 45.5 & 54.8 / 33.2 & 67.7 / 41.3 & 55.3 \\
Sentence entity foils & 71.7 / 36.5 & 65.5 / 31.2 & 66.5 / 34.1 & 55.8 \\
Sentence entity+relation foils & 66.5 / 82.7 & 59.5 / 69.3 & 62.2 / 72.2 & 56.8 \\
CS-CLIP (Qwen) & 73.6 / 79.2 & 70.2 / 63.4 & 68.9 / 63.2 & 57.8 \\
CS-CLIP (Mistral) & 66.9 / 60.4 & 65.5 / 55.0 & 64.5 / 48.0 & 57.3 \\
\bottomrule
\end{tabular*}
\end{table}

\textbf{Matched foils and supervision granularity.}
Table~\ref{tab:foil_controls} compares sentence-level foil training with unit-level
supervision. Adding entity and relation foils at the sentence level raises COCO
(Qwen) relation accuracy from NegCLIP's 45.5\% to 82.7\%. Entity-only sentence
foils reach just 36.5\%, highlighting the contribution of relation foils.
CS-CLIP improves entity accuracy from 66.5\% to 73.6\% and compositional I2T
from 56.8\% to 57.8\% relative to the entity+relation sentence control, while
relation accuracy is slightly lower (79.2\%). Thus, matched foils drive much of
the relation gain; unit supervision improves the balance across capabilities.

\textbf{Generator sensitivity.}
Training with Mistral foils retains gains over CLIP on all three sources and
reaches 57.3\% compositional I2T, compared to 57.8\% for Qwen-trained CS-CLIP.
Its Half-Truth accuracy drops most on relations: 60.4\% versus 79.2\% on COCO
(Qwen). Generator quality therefore matters, although improvements transfer
across generators. Appendix~\ref{app:ablations_full} gives configurations and
additional construction and optimization ablations.

\section{Conclusion}
\label{sec:conclusion}

We study a simple retrieval failure in CLIP-style models: appending one realistic but 
incorrect detail to a correct description can increase similarity, making the half-truth 
rank above the anchor.
To address this, we introduce CS-CLIP, which adds unit-level supervision during 
fine-tuning by parsing captions into entity and relation units and contrasting each 
against a minimally edited foil.
This provides direct compositional pressure while maintaining the standard dual-encoder 
architecture.
CS-CLIP achieves 76.4\% Half-Truth Accuracy on COCO (Qwen), compared to 34.0\%
for CLIP, and retains gains across foil generators and image domains.
Its 57.8\% compositional I2T accuracy shows that these improvements extend to
independently constructed benchmarks.

\textbf{Limitations and future work.}
Our approach relies on text-only LLM parsing, which may introduce artifacts or 
miss visual details not expressed in captions, and fine-tuning on COCO trades 
some zero-shot accuracy for compositional sensitivity.
While CS-CLIP reduces half-truth failures, it does not guarantee factual 
correctness or demographic fairness; models may still reflect biases present 
in training data.
Future work could explore image-side half-truths (adding incorrect visual 
elements to correct images), joint image-text parsing that leverages visual 
grounding, or applying unit-level supervision during large-scale pretraining 
to reduce the zero-shot trade-off.

\begin{ack}
The author B. K. acknowledges support from the Zuse  School ELIZA, which provided
essential funding that enabled their contributions to this work.
\end{ack}




\newpage
\appendix
\makeatletter
\let\appOldSection\section
\renewcommand{\section}{\clearpage\appOldSection}
\makeatother
\clearpage

\section{Sentence-Level Global Objective}
\label{app:global_loss}

This appendix defines the sentence-level contrastive objective $\mathcal{L}_{\mathrm{global}}$ 
from Eq.~\ref{eq:csclip_loss_main} (Section~\ref{sec:method:objective}), and summarizes how
NegCLIP~\cite{yuksekgonul2022when} modifies the denominator with synthetic hard negative 
captions.

We use the scoring notation $\kappa(\mathbf{x},\mathbf{y})=\exp(\mathbf{x}^\top \mathbf{y}/\tau)$ from Section~\ref{sec:method:objective}, where all embeddings are
$\ell_2$-normalized so cosine similarity equals dot product.

\subsection{CLIP global contrastive loss}

Consider a minibatch of $B$ matched image--caption pairs $\{(I_i,T_i)\}_{i=1}^{B}$.
Let $\mathbf{v}_i$ be the normalized image embedding and $\mathbf{t}_i$ be the normalized 
caption embedding.
The image-to-text (I2T) loss is
\begin{equation}
\mathcal{L}_{I\rightarrow T}
= -\frac{1}{B}\sum_{i=1}^{B}\log
\frac{\kappa(\mathbf{v}_i,\mathbf{t}_i)}
{\sum_{j=1}^{B}\kappa(\mathbf{v}_i,\mathbf{t}_j)}.
\label{eq:global_i2t_clip}
\end{equation}
The symmetric text-to-image (T2I) loss is
\begin{equation}
\mathcal{L}_{T\rightarrow I}
= -\frac{1}{B}\sum_{i=1}^{B}\log
\frac{\kappa(\mathbf{v}_i,\mathbf{t}_i)}
{\sum_{j=1}^{B}\kappa(\mathbf{v}_j,\mathbf{t}_i)}.
\label{eq:global_t2i_clip}
\end{equation}
We define the global CLIP objective as
\begin{equation}
\mathcal{L}_{\mathrm{CLIP}}
=\frac{1}{2}\Big(\mathcal{L}_{I\rightarrow T}+\mathcal{L}_{T\rightarrow I}\Big).
\label{eq:global_clip}
\end{equation}

\subsection{NegCLIP-style hard negative captions}

NegCLIP~\cite{yuksekgonul2022when} augments each training example with a 
\emph{shuffled-caption} hard negative.
Given a caption $T_i$, it constructs a foil $\tilde{T}_i$ by shuffling or swapping 
content words (e.g., nouns, adjectives) so that the new sentence reuses nearly the 
same vocabulary but changes the sentence-level meaning.
Let $\tilde{\mathbf{t}}_i$ denote the normalized embedding of the shuffled caption 
$\tilde{T}_i$.

NegCLIP then adds these shuffled negatives to the image-to-text denominator. With a 
minibatch of $B$ pairs, the I2T loss becomes
\begin{equation}
\mathcal{L}_{I\rightarrow T}^{\mathrm{Neg}}
= -\frac{1}{B}\sum_{i=1}^{B}\log
\frac{\kappa(\mathbf{v}_i,\mathbf{t}_i)}
{\sum_{j=1}^{B}\kappa(\mathbf{v}_i,\mathbf{t}_j)\;+\;\sum_{j=1}^{B}\kappa(\mathbf{v}_i,\tilde{\mathbf{t}}_j)}.
\label{eq:global_i2t_negclip}
\end{equation}
That is, each image $\mathbf{v}_i$ is contrasted against all $B$ in-batch captions 
$\{\mathbf{t}_j\}$ plus all $B$ shuffled-caption foils $\{\tilde{\mathbf{t}}_j\}$ 
from the same batch.

The text-to-image term remains unchanged from Eq.~\ref{eq:global_t2i_clip} (i.e., no 
hard negatives are added in the T2I direction).
We define the NegCLIP objective as
\begin{equation}
\mathcal{L}_{\mathrm{NegCLIP}}
=\frac{1}{2}\Big(\mathcal{L}_{I\rightarrow T}^{\mathrm{Neg}}+\mathcal{L}_{T\rightarrow I}\Big).
\label{eq:global_negclip}
\end{equation}

\paragraph{Relation to CS-CLIP.}
Unless otherwise stated, CS-CLIP uses the NegCLIP-style objective 
(Eq.~\ref{eq:global_negclip}) as $\mathcal{L}_{\mathrm{global}}$.
In our ablation studies (Appendix~\ref{app:ablations_full}), we also explore using the 
standard CLIP objective (Eq.~\ref{eq:global_clip}) without hard negatives.
We then add $\lambda_u\,\mathcal{L}_{\mathrm{unit}}$ (Eq.~\ref{eq:csclip_loss_main}) to 
provide supervision on individual caption units and their matched foils.
\section{Unit Parsing and Foil Generation Pipeline}
\label{app:llm_parsing}

This appendix describes the text-only LLM pipeline used to (i) parse captions into 
entity units and relation units, and (ii) generate matched foils via minimal edits.
The same pipeline is used both for CS-CLIP training (Section~\ref{sec:method:units}) 
and for constructing the half-truth diagnostic (Section~\ref{sec:motivation:half_truth}).
The LLM operates on caption text only and never sees images.
For evaluation, this pipeline generates candidates; subsequent image-grounded
checks and manual verification are described in Appendix~\ref{app:half_truth_per_condition}.

\paragraph{Units and foils.}
Given a caption $T$, we extract two types of units:
\begin{itemize}[leftmargin=1.2em, itemsep=0.2em]
  \item \textbf{Entity units} $\mathcal{E}(T)$: noun phrases describing visually grounded 
  entities, including bound attributes and quantifiers (e.g., \emph{``three dogs''}, 
  \emph{``a brown horse''}, \emph{``a man in a blue shirt''}).
  \item \textbf{Relation units} $\mathcal{R}(T)$: directed relations between two entity 
  units, written as $(e_s, p, e_o)$ and rendered as short phrases (e.g., \emph{``person 
  riding horse''}, \emph{``ball in park''}).
\end{itemize}
For each unit, we generate one or more \textbf{matched foils} $\tilde{U}$ that are 
minimally edited (changing exactly one component) and realistic in context, while 
changing the unit's meaning.
Entity foils change either the object category or one attribute/quantifier.
Relation foils change a single relation component (predicate or one argument), or swap 
arguments when the predicate is asymmetric.

\subsection{Model and Inference Configuration}
\label{app:llm_parsing:config}

We use Qwen3-8B-AWQ served with vLLM to produce structured JSON outputs.
Table~\ref{tab:appendix_model_config} summarizes the configuration.

\begin{table}[t]
\centering
\caption{Model and inference configuration for the text-only pipeline.}
\label{tab:appendix_model_config}
\small
\begin{tabular}{@{}ll@{}}
\toprule
\textbf{Parameter} & \textbf{Value} \\
\midrule
Model & \texttt{Qwen3-8B-AWQ} \\
Quantization & AWQ (4-bit) \\
Inference framework & vLLM \\
Batch size & 256 \\
Output format & Structured JSON \\
\bottomrule
\end{tabular}
\end{table}

\subsection{Pipeline Overview}
\label{app:llm_parsing:overview}

For each caption $T$, the pipeline runs three stages:
\begin{enumerate}[leftmargin=1.3em, itemsep=0.2em]
  \item \textbf{Extract units:} produce entity units $\mathcal{E}(T)$ and relation units 
  $\mathcal{R}(T)$.
  \item \textbf{Generate entity foils:} for each $e\in\mathcal{E}(T)$, generate minimally 
  edited foils $\tilde{e}$.
  \item \textbf{Generate relation foils:} for each $r\in\mathcal{R}(T)$, generate minimally 
  edited foils $\tilde{r}$.
\end{enumerate}
We apply lightweight rule-based filters to remove malformed outputs and avoid degenerate 
foils (e.g., near-synonyms or duplicates).

\subsection{Entity and Relation Extraction}
\label{app:llm_parsing:extraction}

This stage parses a caption into entity units and relation units.

\subsubsection{System Prompt (Extraction)}

\begin{tcolorbox}[
  colback=gray!4,
  colframe=gray!55,
  title={\textbf{System Prompt: Entity and Relation Extraction}},
  fonttitle=\small\ttfamily,
  boxrule=0.5pt,
  arc=2pt,
  breakable
]
\small\ttfamily
You extract units for vision-language retrieval.

ENTITY UNITS:
- Output noun phrases for visually grounded entities.
- Keep bound attributes and quantifiers inside the noun phrase.
  Examples: "three dogs", "a brown horse", "a man in a blue shirt".
- Do not output abstract concepts or non-visual phrases.
- Do not duplicate entities (avoid both "brown horse" and "horse").

RELATION UNITS:
- Output directed relations as (subject, predicate, object).
- Subject and object must be chosen from the extracted entity units.
- Use short, visually checkable predicates (spatial, action, possession/association).
- Keep directionality: subject does predicate to object.

Return ONLY valid JSON:
\{
  "entities": ["..."],
  "relations": [
    \{"subject": "...", "predicate": "...", "object": "..."\}
  ]
\}
\end{tcolorbox}

\subsubsection{User Prompt Template (Extraction)}

\begin{tcolorbox}[
  colback=gray!4,
  colframe=gray!55,
  title={\textbf{User Prompt Template}},
  fonttitle=\small\ttfamily,
  boxrule=0.5pt,
  arc=2pt
]
\small\ttfamily
Extract entity units and relation units from this caption:

Caption: "\{caption\}"

Return only valid JSON.
\end{tcolorbox}

\subsubsection{Extraction constraints}
We apply the following constraints:
\begin{itemize}[leftmargin=1.2em, itemsep=0.2em]
  \item \textbf{Attribute binding:} keep modifiers attached to the noun phrase head (e.g., 
  \emph{``three dogs''}, not \emph{``dogs''}).
  \item \textbf{Groundedness:} filter generic placeholders (e.g., \emph{``thing''}) and 
  non-visual phrases.
  \item \textbf{Relation grounding:} require relation arguments to match extracted entity 
  strings exactly.
\end{itemize}

\subsection{Matched Foils for Entity Units}
\label{app:llm_parsing:entity_foils}

For each entity unit, we generate foils that change exactly one component while staying 
realistic.

\subsubsection{System Prompt (Entity Foils)}

\begin{tcolorbox}[
  colback=blue!3,
  colframe=blue!45,
  title={\textbf{System Prompt: Entity Foil Generation}},
  fonttitle=\small\ttfamily,
  boxrule=0.5pt,
  arc=2pt,
  breakable
]
\small\ttfamily
Generate matched foils for an entity noun phrase using minimal edits.

Allowed edits (choose one per foil):
1) Object change: replace the head noun with a similar-category object, keep 
attributes/quantifiers.
   Example: "three dogs" -> "three cats"
2) Attribute/quantifier change: change exactly one attribute or the quantity, keep the 
head noun.
   Example: "a brown horse" -> "a white horse"

Requirements:
- Generate 2-4 foils.
- Each foil must differ by exactly one edit.
- Avoid synonyms (e.g., couch->sofa).
- Keep a similar specificity level.

Return ONLY valid JSON:
\{ "foils": [ \{"text": "...", "edit\_type": "object\_change"\}, ... ] \}
\end{tcolorbox}

\subsubsection{User Prompt Template (Entity Foils)}

\begin{tcolorbox}[
  colback=blue!3,
  colframe=blue!45,
  title={\textbf{User Prompt Template}},
  fonttitle=\small\ttfamily,
  boxrule=0.5pt,
  arc=2pt
]
\small\ttfamily
Generate matched foils for this entity unit.

Entity unit: "\{entity\_unit\}"
Caption context: "\{caption\}"

Return only valid JSON.
\end{tcolorbox}

\subsection{Matched Foils for Relation Units}
\label{app:llm_parsing:relation_foils}

For each relation unit $(e_s, p, e_o)$, we generate foils that change one component 
while keeping arguments grounded.

\subsubsection{System Prompt (Relation Foils)}

\begin{tcolorbox}[
  colback=orange!4,
  colframe=orange!55,
  title={\textbf{System Prompt: Relation Foil Generation}},
  fonttitle=\small\ttfamily,
  boxrule=0.5pt,
  arc=2pt,
  breakable
]
\small\ttfamily
Generate matched foils for a directed visual relation (subject, predicate, object).

Allowed edits (one per foil):
1) Predicate change: replace the predicate with a different plausible predicate.
2) Argument swap: swap subject and object ONLY if the predicate is asymmetric.
3) Argument change: replace ONE argument with a similar-category entity.

Requirements:
- Generate 1-3 foils.
- Keep subject/object strings grounded (use entities from the caption when possible).
- Avoid symmetric swaps (e.g., "next to", "near", "with").
- Keep the text fluent and minimal-edit.

Return ONLY valid JSON:
\{ "foils": [ \{"subject":"...", "predicate":"...", "object":"..."

,"edit\_type":
"predicate\_change"\}, ... ] \}
\end{tcolorbox}

\subsubsection{User Prompt Template (Relation Foils)}

\begin{tcolorbox}[
  colback=orange!4,
  colframe=orange!55,
  title={\textbf{User Prompt Template}},
  fonttitle=\small\ttfamily,
  boxrule=0.5pt,
  arc=2pt
]
\small\ttfamily
Generate matched foils for this relation.

Relation: <\{subject\}, \{predicate\}, \{object\}>
Caption context: "\{caption\}"

Return only valid JSON.
\end{tcolorbox}

\subsection{Post-processing and Filters}
\label{app:llm_parsing:filters}

We apply lightweight filters to remove invalid outputs:
\begin{itemize}[leftmargin=1.2em, itemsep=0.2em]
  \item \textbf{Duplicate removal:} drop foils identical to the original unit or duplicated 
  within the same caption.
  \item \textbf{Near-synonym filter:} drop obvious paraphrases that preserve meaning.
  \item \textbf{Minimal-edit check:} prefer single-component edits; reject multi-edit 
  strings when detected.
  \item \textbf{Swap validity:} block swaps for symmetric predicates (e.g., \emph{next to}, 
  \emph{near}, \emph{with}).
\end{itemize}

\subsection{Usage in Training and Evaluation}
\label{app:llm_parsing:usage}

\paragraph{CS-CLIP training.}
For each image-caption pair $(I,T)$, we sample one unit $U$ from $\mathcal{E}(T)\cup\mathcal{R}(T)$ 
and one matched foil $\tilde{U}$ produced by this pipeline (Section~\ref{sec:method:units}).

\paragraph{Half-truth diagnostic.}
Given an anchor entity unit $A\in\mathcal{E}(T)$, we form a half-truth $A^-$ by appending 
exactly one foil from a different unit (Section~\ref{sec:motivation:half_truth}).

\section{Half-Truth Diagnostic: Construction and Additional Results}
\label{app:half_truth}

\subsection{Construction and Notation}
\label{app:half_truth_construction}

This appendix details how we construct the texts used in the half-truth diagnostic 
(Section~\ref{sec:motivation:half_truth}). The construction follows a simple refinement 
template: start from a caption-supported \emph{anchor} entity description and add 
\emph{exactly one} additional piece of information.

\paragraph{Parsed caption structure.}
For each evaluation pair $(I,T)$, we run our text-only parsing pipeline on caption $T$ 
(Appendix~\ref{app:llm_parsing}) and obtain:
(i) a set of \textbf{entity units} $\mathcal{E}(T)$ (noun phrases with bound 
attributes/quantifiers, e.g., \textit{``red car''}, \textit{``three dogs''}), and
(ii) a set of \textbf{relation units} $\mathcal{R}(T)$ as triplets
$r=\langle e_s,p,e_o\rangle$ where $e_s,e_o\in\mathcal{E}(T)$ and $p$ is a visually 
grounded predicate.
For each unit we also construct matched minimal foils:
$\tilde{\mathcal{E}}(e)$ for entity units and $\tilde{\mathcal{R}}(r)$ for relation units.

\paragraph{Text serialization.}
All comparisons are defined over text strings. We use a single serialization operator 
$\operatorname{txt}(\cdot)$:
\begin{itemize}[leftmargin=1.2em, itemsep=0.2em]
  \item \textbf{Entity:} $\operatorname{txt}(e)$ is the surface string of entity unit $e$.
  \item \textbf{Entity conjunction:} $\operatorname{txt}(e_a,e_b)=\operatorname{txt}(e_a)
  \texttt{ and }\operatorname{txt}(e_b)$.
  \item \textbf{Relation:} for $r=\langle e_s,p,e_o\rangle$, $\operatorname{txt}(r)=
  \operatorname{txt}(e_s)\;p\;\operatorname{txt}(e_o)$.
\end{itemize}

\paragraph{Anchor (short description).}
We sample an anchor entity unit $e_a \in \mathcal{E}(T)$ and define the anchor text
\[
A \;=\; \operatorname{txt}(e_a).
\]
The anchor is \emph{caption-supported} in the sense that it is extracted from $T$.

\paragraph{Truthful vs.\ half-truth additions.}
We consider two ways to refine the anchor by adding one unit:
\begin{itemize}[leftmargin=1.2em, itemsep=0.2em]
  \item A \textbf{truthful addition} uses a caption-consistent unit (another entity unit 
  or a relation unit involving the anchor) extracted from the same caption.
  \item A \textbf{half-truth} uses the same refinement template, but replaces the added 
  unit with a \textbf{matched foil} produced by a minimal edit.
\end{itemize}
By construction, the anchor and its truthful/half-truth refinements share most tokens, 
isolating whether the model verifies the \emph{added unit}.

\subsubsection{Entity addition: adding one more entity}
\label{app:half_truth_entity_construction}

\paragraph{Truthful entity addition and half-truth.}
We sample a second entity unit $e^{+}\in\mathcal{E}(T)\setminus\{e_a\}$ and form the 
truthful refinement
\[
A^{+}_{e} \;=\; \operatorname{txt}(e_a,e^{+})
\;=\;
\operatorname{txt}(e_a)\texttt{ and }\operatorname{txt}(e^{+}).
\]
We then keep the anchor fixed and replace only the added entity unit with a matched foil:
\[
A^{-}_{e} \;=\; \operatorname{txt}(e_a,e^{-}),
\qquad e^{-} \in \tilde{\mathcal{E}}(e^{+}).
\]

\paragraph{Entity foil types (\textbf{+Obj}, \textbf{+Attr}).}
In the verified evaluation set, we group entity half-truths by how the added entity unit is corrupted:
\begin{itemize}[leftmargin=1.2em, itemsep=0.2em]
  \item \textbf{+Obj (wrong object):} replace the head noun of $e^{+}$ while keeping 
  modifiers fluent (e.g., \textit{``red car''}$\rightarrow$\textit{``red truck''}).
  \item \textbf{+Attr (wrong attribute):} replace an attribute/modifier while keeping 
  the head noun (e.g., \textit{``red car''}$\rightarrow$\textit{``blue car''}).
\end{itemize}

\subsubsection{Relation addition: adding one relation involving the anchor}
\label{app:half_truth_relation_construction}

\paragraph{Truthful relation addition and half-truth.}
We sample a relation unit $r^{+}\in\mathcal{R}(T)$ in which the anchor entity participates 
(as subject or object).
We form the truthful refinement by appending this relation to the anchor:
\[
A^{+}_{r} \;=\; \operatorname{txt}(e_a)\texttt{ and }\operatorname{txt}(r^{+}).
\]
We then form a half-truth by substituting a matched foil relation:
\[
A^{-}_{r} \;=\; \operatorname{txt}(e_a)\texttt{ and }\operatorname{txt}(r^{-}),
\qquad r^{-}\in\tilde{\mathcal{R}}(r^{+}).
\]

\paragraph{Relation foil types (\textbf{Ant}, \textbf{Swap}).}
In the verified evaluation set, we group relation half-truths by whether the predicate or the argument order is perturbed:
\begin{itemize}[leftmargin=1.2em, itemsep=0.2em]
  \item \textbf{Ant (predicate change):} replace $p$ with an opposing relation
  (e.g., \textit{on}$\rightarrow$\textit{under}, \textit{in front of}$\rightarrow$
  \textit{behind}).
  \item \textbf{Swap (role swap):} swap subject and object
  ($\langle e_s,p,e_o\rangle \rightarrow \langle e_o,p,e_s\rangle$),
  preserving the predicate but changing directionality.
\end{itemize}

\begin{figure*}[t]
  \centering
  \includegraphics[width=\textwidth]{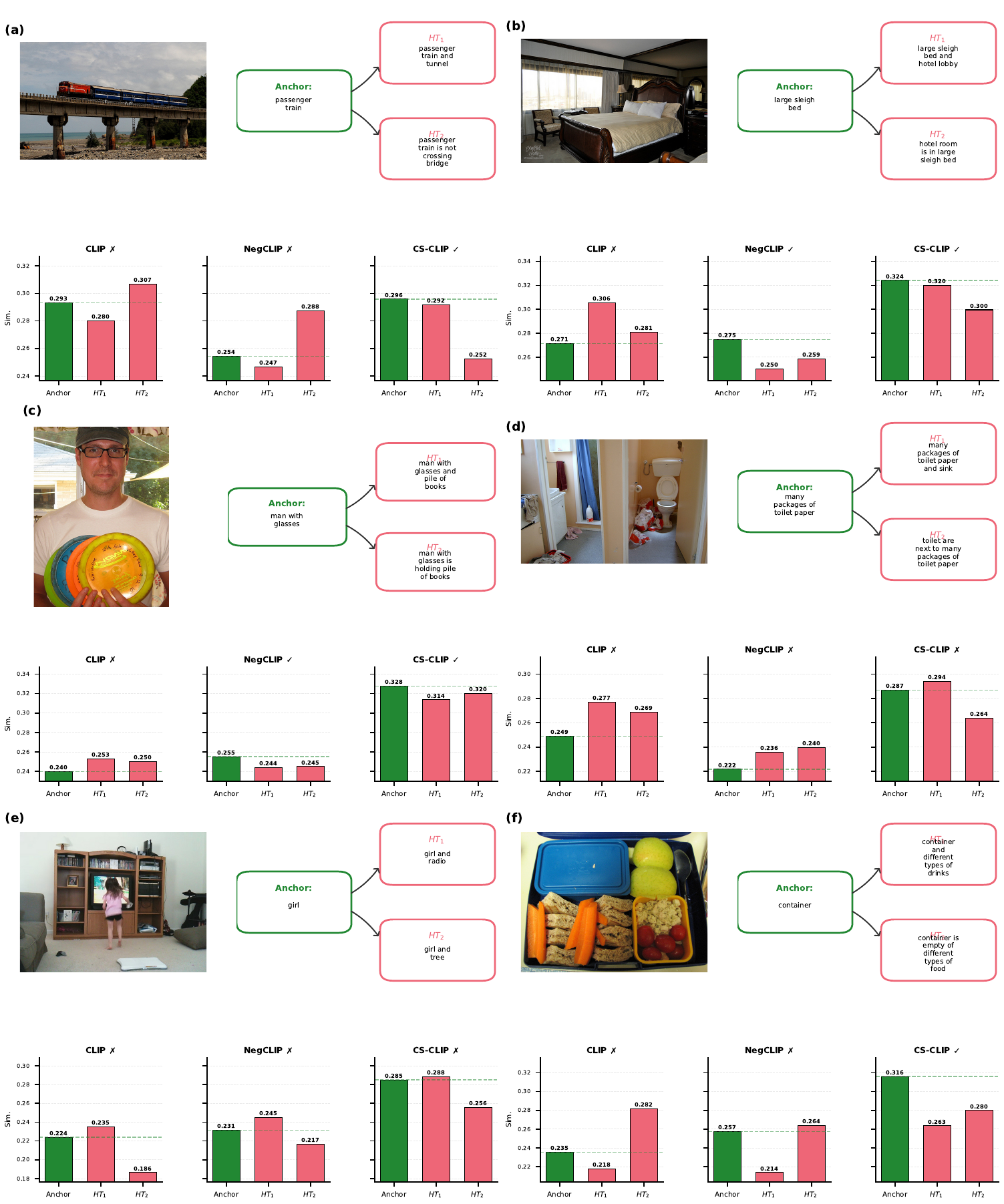}
  \caption{\textbf{Additional half-truth examples with model scores.}
  Multiple anchors and their half-truth refinements with similarity scores from CLIP, 
  NegCLIP, and CS-CLIP. Some baselines assign comparable or higher similarity to 
  incorrect refinements, while CS-CLIP more consistently lowers similarity for half-truths.}
  \label{fig:app_ht_multi_samples}
\end{figure*}

\paragraph{Why edits are minimal.}
All half-truth variants are generated by a single localized perturbation so the anchor 
and its refinements share most tokens.
This keeps the comparison focused on verifying the added unit rather than topic drift.
Figure~\ref{fig:app_ht_multi_samples} provides additional qualitative examples comparing 
CLIP, NegCLIP, and CS-CLIP similarity scores across multiple half-truth examples.

\subsection{Evaluation Sources and Verification}
\label{app:half_truth_per_condition}
The evaluation covers two foil generators and two image domains. Entity foils
change an object or attribute; relation foils change a predicate or swap its arguments.
We first check candidate foils against their images with Qwen3-VL-30B.
We then use Claude Opus 5 to review the candidates and propose replacements
within the same foil category. Finally, we manually inspect the candidates,
including proposed replacements, and retain additions judged incorrect for the image.
Candidates judged correct or ambiguous are discarded.

\begin{table}[t]
\centering
\caption{Manually verified evaluation comparisons by source and foil type.}
\label{tab:half_truth_per_condition_coco}
\small
\begin{tabular}{lrrrrr}
\toprule
Source & Object & Attribute & Predicate & Swap & Total \\
\midrule
COCO (Qwen) & 195 & 59 & 167 & 88 & 509 \\
COCO (Mistral) & 177 & 75 & 159 & 43 & 454 \\
CC3M (Qwen) & 173 & 78 & 157 & 66 & 474 \\
\bottomrule
\end{tabular}
\end{table}

\paragraph{Training-foil quality.}
Training foils are automatically generated and filtered, without manual verification
at scale. An image-grounded Qwen3-VL-30B audit of 1,000 foils flags 246 as
accidentally true (24.6\%): 132/500 entity foils (26.4\%) and 114/500 relation
foils (22.8\%). This is a model-based estimate. On 816 independently human-labeled
verification decisions, the verifier agrees with annotators on 80.3\% of entity
cases and 55.8\% of relation cases. These errors motivate manual evaluation checks;
the training results show gains despite estimated foil noise, rather than establishing
that false negatives have no effect.

\subsection{Correct Ordering Rate}
\label{app:half_truth_ordering}
To test whether a model distinguishes correct additions from incorrect ones,
we measure Correct Ordering Rate (COR), $\Pr[s(I,A^{+})>s(I,A)>s(I,A^{-})]$,
where $A^{+}$ is a correct refinement of the anchor. All three descriptions
refer to the same image.

\begin{table*}[t]
\centering
\scriptsize
\setlength{\tabcolsep}{3pt}
\renewcommand{\arraystretch}{1.08}
\caption{\textbf{Correct Ordering Rate across three evaluation sources.} COR measures $\Pr[s(I,A^{+})>s(I,A)>s(I,A^{-})]$.
We report percentages on 509 COCO (Qwen), 454 COCO (Mistral), and 474 CC3M (Qwen) manually verified comparisons. All is the sample-weighted entity/relation average. FSC-CLIP is trained on CC3M; the other fine-tuned models use COCO. Higher is better.}
\label{tab:verified_ft_st_ht_ordering}
\begin{tabular*}{\textwidth}{@{\extracolsep{\fill}}l|ccc|ccc|ccc@{}}
\toprule
& \multicolumn{3}{c|}{\textbf{COCO (Qwen)}} & \multicolumn{3}{c|}{\textbf{COCO (Mistral)}} & \multicolumn{3}{c}{\textbf{CC3M (Qwen)}} \\
\textbf{Model} & All & Entity & Relation & All & Entity & Relation & All & Entity & Relation \\
\midrule
CLIP & 18.9 & 26.0 & 11.8 & 19.2 & 26.6 & 9.9 & 24.5 & 34.7 & 13.0 \\
NegCLIP & 32.2 & 40.6 & 23.9 & 25.8 & 36.1 & 12.9 & 29.1 & 41.4 & 15.2 \\
DeGLA & 32.4 & 40.6 & 24.3 & 28.2 & 38.1 & 15.8 & 31.4 & 41.8 & 19.7 \\
ReadCLIP & 35.0 & 38.2 & 31.8 & 31.7 & 38.5 & 23.3 & 30.2 & 35.1 & 24.7 \\
FSC-CLIP (CC3M) & 24.2 & 36.2 & 12.2 & 25.6 & 35.3 & 13.4 & 26.8 & 39.8 & 12.1 \\
\rowcolor{gray!10}CS-CLIP & \textbf{58.2} & \textbf{56.3} & \textbf{60.0} & \textbf{50.4} & \textbf{50.0} & \textbf{51.0} & \textbf{42.2} & \textbf{43.8} & \textbf{40.4} \\
\bottomrule
\end{tabular*}
\end{table*}

\begin{figure}[t]
  \centering
  \includegraphics[width=\linewidth]{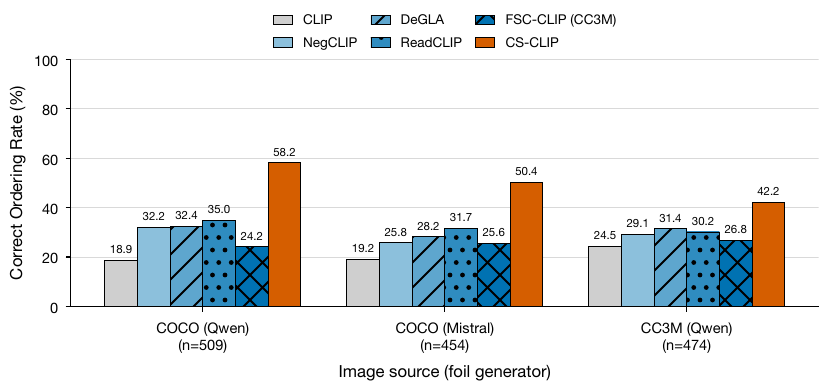}
  \caption{\textbf{Correct ordering across evaluation sources.}
  COR on the same comparisons as Table~\ref{tab:halftruth_overall_and_split}.
  Each bar reports one checkpoint evaluation; uncertainty is not shown.}
  \label{fig:verified_ft_st_ht_ordering}
\end{figure}

CS-CLIP achieves the highest COR among the displayed models: 58.2\% on COCO
(Qwen), 50.4\% on COCO (Mistral), and 42.2\% on CC3M (Qwen).
On COCO (Qwen), its relational COR is 60.0\%, compared to 11.8\% for CLIP and
23.9\% for NegCLIP. Requiring both inequalities prevents a uniform preference
for shorter captions from satisfying the metric.

\section{Compositional Benchmark Suite and Capability Taxonomy}
\label{app:capability_mapping}

We evaluate compositional robustness using a heterogeneous suite of 16 benchmarks that test 
entity content, relational structure, attribute binding, and linguistic phenomena through 
controlled perturbations.
To enable fine-grained analysis beyond aggregate scores, we organize benchmark subsets into 
a shared capability taxonomy with four top-level categories and ten sub-capabilities.

\subsection{Benchmark Suite}
\label{app:benchmark_suite}

Table~\ref{tab:comp_benchmark_overview} summarizes our compositional benchmark suite.
Most benchmarks provide \emph{foil captions} that preserve topical content while changing a 
targeted semantic factor (e.g., swapping attributes, changing predicates, or altering word order).
Several benchmarks additionally provide \emph{foil images} for stricter cross-modal consistency 
checks.

\begin{table}[t]
\centering
\caption{\textbf{Compositional benchmark suite.}
Text: foil captions provided; Img: foil images provided.}
\label{tab:comp_benchmark_overview}
\scriptsize
\setlength{\tabcolsep}{3pt}
\renewcommand{\arraystretch}{1.08}
\begin{tabular}{@{}llccp{0.46\linewidth}@{}}
\toprule
\textbf{Benchmark} & \textbf{Cite} & \textbf{Text} & \textbf{Img} & \textbf{Primary phenomenon} \\
\midrule
ARO & \cite{yuksekgonul2022when} & \cmark & & Binding, relations, word order \\
BLA & \cite{chen2023bla} & \cmark & & Syntax and word order \\
COCO-CF & \cite{le2023cococf} & \cmark & \cmark & Counterfactual edits \\
COLA & \cite{ray2024cola} & \cmark & \cmark & Multi-object binding \\
ColorFoil & \cite{samin2025colorfoil} & \cmark & & Color attributes \\
ColorSwap & \cite{burapacheep2024colorswap} & \cmark & & Attribute binding \\
MMVP & \cite{tong2024eyes} & \cmark & \cmark & Visual attributes, spatial \\
NegBench & \cite{alhamoud2025vision} & \cmark & & Negation and absence \\
SPEC & \cite{peng2024spec} & \cmark & \cmark & Synthetic controlled scenes \\
SugarCrepe & \cite{hsieh2023sugarcrepe} & \cmark & & Fluent minimal edits \\
SugarCrepe++ & \cite{dumpala2024sugarcrepepp} & \cmark & & Paraphrase invariance \\
VALSE & \cite{parcalabescu2022valse} & \cmark & & Existence, counting, coreference \\
VisMin & \cite{awal2024vismin} & \cmark & \cmark & Minimal visual changes \\
What'sUp & \cite{kamath2023whats} & \cmark & \cmark & Spatial relations \\
Winoground & \cite{thrush2022winoground} & \cmark & \cmark & Role sensitivity \\
VL-CheckList & \cite{zhao2022vlchecklist} & \cmark & & Objects, attributes, relations \\
\bottomrule
\end{tabular}
\end{table}

\subsection{Capability Taxonomy}
\label{app:capability_taxonomy}

To analyze which compositional skills are improved by different methods, we categorize 
benchmark subsets into a shared taxonomy.
For each subset, we assign a top-level \textbf{category} and a \textbf{sub-capability} based 
on the minimal semantic change needed to distinguish correct captions from foils.
This enables macro-averaged reporting at the category level and fine-grained analysis at the 
sub-capability level.

\paragraph{Top-level categories.}
We define four categories (Table~\ref{tab:capability_taxonomy}):
\begin{itemize}[leftmargin=1.2em, itemsep=0.2em]
  \item \textbf{Entity Content:} Recognizing objects, attributes, quantities, and existence.
  \item \textbf{Relational Structure:} Understanding predicates and role assignments in relations.
  \item \textbf{Binding:} Associating attributes with the correct objects.
  \item \textbf{Linguistic:} Handling syntax, coreference, and negation.
\end{itemize}
Category scores are computed by macro-averaging subset accuracies within each category.

\paragraph{Sub-capabilities.}
We define ten sub-capabilities for fine-grained analysis:
\begin{itemize}[leftmargin=1.2em, itemsep=0.2em]
  \item \textbf{Object Recognition:} Correct object category (e.g., \emph{dog} vs.\ \emph{cat}).
  \item \textbf{Attribute Recognition:} Correct attribute (e.g., \emph{red} vs.\ \emph{blue}).
  \item \textbf{Existence:} Whether an object is present.
  \item \textbf{Counting:} Correct number of objects (e.g., \emph{four} vs.\ \emph{three horses}).
  \item \textbf{Predicate Sensitivity:} Correct relation predicate (e.g., \emph{left of} vs.\ 
  \emph{right of}).
  \item \textbf{Role Sensitivity:} Which entity plays which role (e.g., \emph{dog chasing cat} 
  vs.\ \emph{cat chasing dog}).
  \item \textbf{Attribute Binding:} Which attribute binds to which object (e.g., \emph{red car 
  and blue cat} vs.\ \emph{blue car and red cat}).
  \item \textbf{Syntax:} Word order changes meaning.
  \item \textbf{Coreference:} Resolving references across phrases.
  \item \textbf{Negation:} Understanding negation (e.g., \emph{has A but not B}).
\end{itemize}

\begin{table}[t]
\centering
\caption{\textbf{Capability taxonomy summary.}}
\label{tab:capability_taxonomy}
\small
\setlength{\tabcolsep}{6pt}
\renewcommand{\arraystretch}{1.12}
\begin{tabular}{@{}ll@{}}
\toprule
\textbf{Category} & \textbf{Sub-capabilities} \\
\midrule
Entity Content & Object Recognition, Attribute Recognition, Counting, Existence \\
Relational Structure & Predicate Sensitivity, Role Sensitivity \\
Binding & Attribute Binding \\
Linguistic & Syntax, Coreference, Negation \\
\bottomrule
\end{tabular}
\end{table}

\subsection{Benchmark Subset Mapping}
\label{app:subset_capability_mapping}

Table~\ref{tab:subset_capability_map} provides the complete mapping from benchmark subsets 
to our taxonomy, organized by category.

\vspace{3mm}
\small
\setlength{\LTpre}{0pt}
\setlength{\LTpost}{0pt}
\setlength{\tabcolsep}{4pt}
\renewcommand{\arraystretch}{1.08}
\begin{longtable}{@{}llll@{}}
\caption{\textbf{Benchmark subset to capability mapping.}}
\label{tab:subset_capability_map}\\
\toprule
\textbf{Benchmark} & \textbf{Subset} & \textbf{Category} & \textbf{Sub-capability} \\
\midrule
\endfirsthead
\toprule
\textbf{Benchmark} & \textbf{Subset} & \textbf{Category} & \textbf{Sub-capability} \\
\midrule
\endhead
\midrule
\multicolumn{4}{r}{\textit{(continued)}}\\
\endfoot
\bottomrule
\endlastfoot

\multicolumn{4}{@{}l}{\textit{\textbf{Entity Content}}} \\
\midrule
SugarCrepe & add\_att & Entity Content & Attribute Recognition \\
SugarCrepe & replace\_att & Entity Content & Attribute Recognition \\
SugarCrepe & add\_obj & Entity Content & Object Recognition \\
SugarCrepe & replace\_obj & Entity Content & Object Recognition \\
SugarCrepe++ & replace\_attribute & Entity Content & Attribute Recognition \\
SugarCrepe++ & replace\_object & Entity Content & Object Recognition \\
VisMin & object & Entity Content & Object Recognition \\
VisMin & attribute & Entity Content & Attribute Recognition \\
VisMin & counting & Entity Content & Counting \\
SPEC & count & Entity Content & Counting \\
SPEC & absolute\_size & Entity Content & Attribute Recognition \\
SPEC & existence & Entity Content & Existence \\
VL-CheckList & attr\_color & Entity Content & Attribute Recognition \\
VL-CheckList & attr\_material & Entity Content & Attribute Recognition \\
VL-CheckList & attr\_size & Entity Content & Attribute Recognition \\
VL-CheckList & attr\_state & Entity Content & Attribute Recognition \\
VL-CheckList & attr\_action & Entity Content & Attribute Recognition \\
VL-CheckList & obj\_location & Entity Content & Object Recognition \\
VL-CheckList & obj\_size & Entity Content & Object Recognition \\
VALSE & existence & Entity Content & Existence \\
VALSE & counting & Entity Content & Counting \\
VALSE & plurals & Entity Content & Counting \\
ColorFoil &  & Entity Content & Attribute Recognition \\
COCO-CF &  & Entity Content & Object Recognition \\
MMVP & Color & Entity Content & Attribute Recognition \\
MMVP & Presence & Entity Content & Existence \\
MMVP & Quantity & Entity Content & Counting \\
MMVP & Structural\ Char. & Entity Content & Attribute Recognition \\
\midrule

\multicolumn{4}{@{}l}{\textit{\textbf{Relational Structure}}} \\
\midrule
ARO & VG\_Relation & Relational Structure & Predicate Sensitivity \\
SugarCrepe & replace\_rel & Relational Structure & Predicate Sensitivity \\
SugarCrepe & swap\_obj & Relational Structure & Predicate Sensitivity \\
SugarCrepe++ & replace\_relation & Relational Structure & Predicate Sensitivity \\
SugarCrepe++ & swap\_object & Relational Structure & Predicate Sensitivity \\
VisMin & relation & Relational Structure & Predicate Sensitivity \\
SPEC & relative\_spatial & Relational Structure & Role Sensitivity \\
SPEC & absolute\_position & Relational Structure & Role Sensitivity \\
SPEC & relative\_size & Relational Structure & Role Sensitivity \\
What'sUp & A & Relational Structure & Role Sensitivity \\
What'sUp & B & Relational Structure & Role Sensitivity \\
What'sUp & VG-One & Relational Structure & Role Sensitivity \\
What'sUp & VG-Two & Relational Structure & Role Sensitivity \\
What'sUp & COCO-One & Relational Structure & Role Sensitivity \\
What'sUp & COCO-Two & Relational Structure & Role Sensitivity \\
VL-CheckList & rel\_action & Relational Structure & Role Sensitivity \\
VL-CheckList & rel\_spatial & Relational Structure & Role Sensitivity \\
VALSE & relations & Relational Structure & Predicate Sensitivity \\
VALSE & actions & Relational Structure & Role Sensitivity \\
MMVP & Orientation & Relational Structure & Role Sensitivity \\
MMVP & Spatial & Relational Structure & Role Sensitivity \\
MMVP & State & Relational Structure & Role Sensitivity \\
Winoground &  & Relational Structure & Role Sensitivity \\
\midrule

\multicolumn{4}{@{}l}{\textit{\textbf{Binding}}} \\
\midrule
ARO & VG\_Attribution & Binding & Attribute Binding \\
SugarCrepe & swap\_att & Binding & Attribute Binding \\
SugarCrepe++ & swap\_attribute & Binding & Attribute Binding \\
ColorSwap &  & Binding & Attribute Binding \\
COLA & multi\_object & Binding & Attribute Binding \\
\midrule

\multicolumn{4}{@{}l}{\textit{\textbf{Linguistic}}} \\
\midrule
ARO & COCO\_Order & Linguistic & Syntax \\
ARO & Flickr30k\_Order & Linguistic & Syntax \\
VALSE & coreference & Linguistic & Coreference \\
VALSE & noun\ phrases & Linguistic & Syntax \\
BLA & ap & Linguistic & Syntax \\
BLA & co & Linguistic & Syntax \\
BLA & rc & Linguistic & Syntax \\
NegBench & MSRVTT & Linguistic & Negation \\
NegBench & VOC2007 & Linguistic & Negation \\
NegBench & COCO & Linguistic & Negation \\

\end{longtable}
\section{Compositional Benchmark Suite: Additional Analyses}
\label{app:compositional}

This appendix provides detailed breakdowns for the compositional evaluation in 
Section~\ref{sec:exp:compositionality}.
In the main paper, we report aggregate suite-level performance showing CS-CLIP achieves 
57.8\% average I2T accuracy across 16 benchmarks (Figure~\ref{fig:comp_dataset_delta}).
Here we provide: (i) per-benchmark results (Table~\ref{tab:summary_compositional}), 
(ii) per-subset visualizations (Figures~\ref{fig:subset_heatmap_delta_part1}--\ref{fig:subset_heatmap_delta_part2}),
(iii) per-capability breakdowns (Table~\ref{tab:capability_sub}), and (iv) downstream
retrieval results and the zero-shot classification trade-off (Section~\ref{app:downstream}).

All analyses use the same models and evaluation protocol as in the main paper.
The capability taxonomy and benchmark subset mapping are provided in 
Appendix~\ref{app:capability_mapping}.

\subsection{Evaluation Protocol and Metrics}
\label{app:compositional:metrics}

Compositional benchmarks present \emph{contrast sets} where one item is correct and one 
or more \emph{foils} differ by a minimal semantic change (e.g., swapping roles, changing 
an attribute, or altering word order). 
Let $s(I,T)$ denote the model's image--text similarity score.

\paragraph{Image-to-Text (I2T) accuracy.}
For each image $I_i$, the benchmark provides a candidate caption set
$\mathcal{T}_i=\{T_{i,1},\ldots,T_{i,m_i}\}$ with a single correct caption $T_i^{\star}\in\mathcal{T}_i$.
I2T accuracy is the fraction of cases where the correct caption ranks above all foils:
\begin{equation}
\mathrm{Acc}_{\mathrm{I2T}}
=
\frac{1}{N}\sum_{i=1}^{N}
\mathbb{I}\Big[
s(I_i,T_i^{\star}) \;>\; \max_{T \in \mathcal{T}_i \setminus \{T_i^{\star}\}} s(I_i,T)
\Big].
\label{eq:i2t_acc}
\end{equation}
When $\mathcal{T}_i$ contains exactly one foil, this reduces to a pairwise win-rate.

\paragraph{Text-to-Image (T2I) accuracy.}
For each text query $T_i$, the benchmark provides a candidate image set
$\mathcal{I}_i=\{I_{i,1},\ldots,I_{i,n_i}\}$ with a single correct image
$I_i^{\star}\in\mathcal{I}_i$.
T2I accuracy is the fraction of cases where the correct image ranks above all foils:
\begin{equation}
\mathrm{Acc}_{\mathrm{T2I}}
=
\frac{1}{N}\sum_{i=1}^{N}
\mathbb{I}\Big[
s(I_i^{\star},T_i) \;>\; \max_{I \in \mathcal{I}_i \setminus \{I_i^{\star}\}} s(I,T_i)
\Big].
\label{eq:t2i_acc}
\end{equation}

\paragraph{Group accuracy.}
Some benchmarks provide groups consisting of $n$ matched image-caption pairs
$\{(I_{i,1},T_{i,1}),\ldots,(I_{i,n},T_{i,n})\}$.
Group Accuracy is the fraction of groups where the model assigns the highest similarity
to the \emph{matched} pair over all mismatched pairs, in both directions:
\begin{equation}
\mathrm{Acc}_{\mathrm{Grp}}
=
\frac{1}{N_g}\sum_{i=1}^{N_g}
\mathbb{I}\Big[
\forall j\in\{1,\ldots,n\}:\;
s(I_{i,j},T_{i,j})
>
\max\Big(
\max_{k\neq j} s(I_{i,k},T_{i,j}),
\max_{\ell\neq j} s(I_{i,j},T_{i,\ell})
\Big)
\Big].
\label{eq:grp_acc}
\end{equation}

\paragraph{Averaging.}
We report per-benchmark accuracies and suite-level averages computed as the unweighted mean 
of benchmark-level accuracies.
For capability-wise analysis, we compute accuracies for each benchmark subset, then 
macro-average within each capability category (Appendix~\ref{app:capability_taxonomy}).
CS-CLIP point estimates use the best-performing run; per-benchmark $\pm$ values denote standard deviations across five runs.
All other models are reported with their single published checkpoint.

\subsection{Per-Benchmark Results}
\label{app:compositional:per_benchmark}

\begin{table}[t]
  \centering
  \scriptsize
  \setlength{\tabcolsep}{2pt}
  \renewcommand{\arraystretch}{1.05}
  \caption{\textbf{Compositional benchmark results (I2T accuracy, \%).}
  Each benchmark averaged over all subsets. Rows grouped by training dataset and sorted 
  by overall average within each group.
  SC = SugarCrepe, SC++ = Sugar++, VLC = VL-Checklist.
  \textbf{Bold}: best; \underline{underline}: second best.}
  \label{tab:summary_compositional}
  \resizebox{\textwidth}{!}{
  \begin{tabular}{lccccccccccccccccc}
    \toprule
    Model & ARO & BLA & COCO-CF & COLA & ColorFoil & ColorSwap & WhatsUp & MMVP & NegBench & SPEC & SC & SC++ & VALSE & VLC & VisMin & Winoground & Avg \\
    \midrule
    
    \multicolumn{18}{@{}l}{\textit{Pre-training variants (zero-shot)}} \\
    \midrule
    CLIP & 48.5 & 49.4 & 74.0 & \underline{41.9} & 84.2 & \underline{60.7} & 41.4 & \underline{13.3} & 30.6 & 32.7 & 76.8 & 59.7 & 65.4 & 67.1 & 57.9 & 29.8 & 52.1 \\
    TripletCLIP & 37.9 & 49.7 & 47.6 & 24.3 & 75.5 & 55.7 & 41.6 & 10.4 & 27.3 & 30.5 & 74.0 & 55.4 & 59.8 & 66.2 & 35.8 & 22.2 & 44.6 \\
    LaCLIP & 39.4 & 50.2 & 41.0 & 21.9 & 69.1 & 56.7 & 41.8 & \textbf{15.6} & 25.5 & 29.2 & 63.5 & 46.0 & 57.3 & 67.6 & 33.7 & 26.2 & 42.8 \\
    \midrule
    
    \multicolumn{18}{@{}l}{\textit{Fine-tuned on COCO (direct comparison)}} \\
    \midrule
    \rowcolor{gray!10}\textbf{CS-CLIP (Ours)} & \underline{86.9}\scalebox{0.6}{$\pm$0.2} & 50.2\scalebox{0.6}{$\pm$0.4} & \underline{78.2}\scalebox{0.6}{$\pm$0.2} & 41.0\scalebox{0.6}{$\pm$2.2} & 90.5\scalebox{0.6}{$\pm$0.8} & 59.0\scalebox{0.6}{$\pm$1.2} & \underline{43.5}\scalebox{0.6}{$\pm$0.6} & \underline{13.3}\scalebox{0.6}{$\pm$1.9} & 32.8\scalebox{0.6}{$\pm$1.2} & \underline{36.1}\scalebox{0.6}{$\pm$0.3} & 82.2\scalebox{0.6}{$\pm$0.5} & 67.4\scalebox{0.6}{$\pm$0.3} & 74.8\scalebox{0.6}{$\pm$0.3} & \textbf{79.2}\scalebox{0.6}{$\pm$0.2} & \underline{60.5}\scalebox{0.6}{$\pm$0.2} & 29.8\scalebox{0.6}{$\pm$1.0} & \textbf{57.8} \\
    FSC-CLIP & 86.0 & 49.3 & 77.5 & \textbf{44.8} & 90.8 & 56.0 & 41.2 & 12.6 & 26.0 & 35.3 & 85.2 & 67.5 & 75.7 & 76.8 & \textbf{61.3} & \textbf{32.8} & \underline{57.4} \\
    ReadCLIP & \textbf{90.0} & 50.0 & 73.9 & 31.4 & \underline{91.9} & \textbf{62.0} & \textbf{44.0} & 11.9 & 27.5 & 34.6 & \underline{87.0} & 69.8 & 76.6 & 75.0 & 60.5 & 24.2 & 56.9 \\
    DeGLA & \underline{86.9} & \underline{50.3} & 75.7 & 32.4 & 91.6 & 57.7 & 42.3 & 4.4 & 30.7 & \textbf{36.2} & \textbf{89.2} & 62.8 & \textbf{77.3} & \underline{78.1} & 59.0 & 24.5 & 56.2 \\
    NegCLIP & 82.4 & 49.8 & 76.4 & 32.9 & 87.5 & 55.3 & 42.6 & 10.4 & 28.6 & 35.6 & 82.5 & 62.7 & 74.7 & 74.0 & 59.6 & 30.5 & 55.3 \\
    LabCLIP & 78.9 & 49.5 & 75.8 & 32.9 & 88.0 & 57.7 & 41.9 & 11.1 & 31.2 & 33.5 & 79.9 & 61.0 & 72.0 & 73.9 & 54.8 & 24.5 & 54.2 \\
    CE-CLIP & 76.8 & 50.1 & 69.2 & 24.8 & \textbf{93.0} & 60.3 & 41.8 & 8.9 & 35.7 & 34.3 & 85.9 & 55.6 & \underline{76.8} & 77.7 & 59.4 & 19.8 & 54.4 \\
    DAC (LLM) & 64.9 & 49.0 & 72.2 & 35.7 & 84.7 & 56.0 & 42.6 & 11.1 & 35.9 & 33.0 & 75.6 & 59.0 & 66.7 & 71.8 & 51.5 & 25.5 & 52.2 \\
    DAC (SAM) & 62.2 & 48.8 & 72.8 & 34.3 & 84.8 & 54.7 & 42.8 & 8.9 & \textbf{37.2} & 32.2 & 74.4 & 59.5 & 65.8 & 71.4 & 51.2 & 26.5 & 51.7 \\
    \midrule
    
    \multicolumn{18}{@{}l}{\textit{Fine-tuned on larger/other datasets (CC3M, LAION, RedCaps)}} \\
    \midrule
    FSC-CLIP (CC3M) & 80.4 & 49.1 & 73.7 & 37.1 & 91.1 & 56.3 & 42.8 & 8.9 & 27.0 & 35.7 & 85.2 & 59.7 & 75.1 & 78.0 & 59.0 & 29.2 & 55.5 \\
    CLoVe & 81.7 & \textbf{52.1} & 73.3 & 35.7 & 88.3 & 58.0 & 41.1 & \underline{13.3} & 25.2 & 32.0 & 84.1 & 51.1 & 72.4 & 72.9 & 53.5 & 22.5 & 53.6 \\
    FSC-CLIP (L+C) & 83.9 & \underline{50.3} & 75.8 & 30.0 & 89.8 & 55.0 & 42.8 & 9.6 & 29.1 & 34.3 & 83.4 & 59.9 & 74.0 & 76.1 & 57.8 & 27.8 & 55.0 \\
    CLIC (RedCaps) & 66.8 & 49.4 & 77.4 & 36.2 & 84.9 & 60.0 & 39.9 & 10.4 & 32.4 & 32.6 & 81.8 & \textbf{70.3} & 69.9 & 74.0 & 54.2 & \underline{32.2} & 54.5 \\
    CLIC (LAION) & 67.8 & 49.5 & 78.0 & 34.3 & 84.6 & 58.3 & 40.0 & 10.4 & 30.3 & 33.0 & 81.2 & \underline{69.9} & 70.0 & 75.3 & 55.4 & 31.5 & 54.3 \\
    CON-CLIP & 41.2 & 49.0 & \textbf{78.8} & 36.2 & 83.0 & 56.3 & 41.2 & 8.9 & 34.8 & 33.5 & 76.5 & 60.6 & 66.8 & 68.0 & 54.9 & 29.0 & 51.2 \\
    
TSVLC & 61.1 & 48.2 & 73.2 & 33.3 & 83.9 & 54.3 & 40.6 & 11.9 & \underline{36.5} & 32.7 & 72.4 & 60.5 & 66.9 & 71.4 & 51.8 & 26.8 & 51.6 \\
    \bottomrule
  \end{tabular}
  }
\end{table}

\paragraph{Analysis.}
Table~\ref{tab:summary_compositional} reveals several patterns.
First, fine-tuning on COCO consistently improves compositional robustness: the top-performing 
models (CS-CLIP, FSC-CLIP, ReadCLIP, DeGLA) all fine-tune on COCO and achieve 56--58\% 
average accuracy, compared to 52\% for zero-shot CLIP.
Several compositional benchmarks draw images from COCO or Visual Genome,
including ARO, What'sUp, VALSE, VL-CheckList, and SugarCrepe.
The benchmarks use independently constructed evaluation tasks, but their image sources
are not all independent of the training corpus.
This shared provenance should be considered when interpreting the gains;
Figure~\ref{fig:comp_dataset_delta} alone does not establish whether they reflect
source-specific adaptation or broader compositional improvements.

Second, CS-CLIP achieves the highest overall average (57.8\%) while excelling on benchmarks 
that require tight binding (ARO 86.9\%, VL-CheckList 79.2\%) and spatial reasoning 
(What'sUp 43.5\%, SPEC 36.1\%).
Third, performance varies substantially across benchmarks: models generally perform well on 
color attributes (ColorFoil 84--93\%) but struggle on spatial relations (What'sUp 39--44\%, 
MMVP 4--16\%) and group-level consistency (Winoground 20--33\%).
Finally, training on larger datasets (CC3M, LAION) does not consistently improve 
compositional robustness compared to COCO fine-tuning, suggesting dataset quality and 
alignment signal matter more than scale alone.

\subsection{Per-Subset Visualization}
\label{app:compositional:per_subset}
\begin{figure*}[t]
  \centering
  \includegraphics[width=0.5\textwidth]{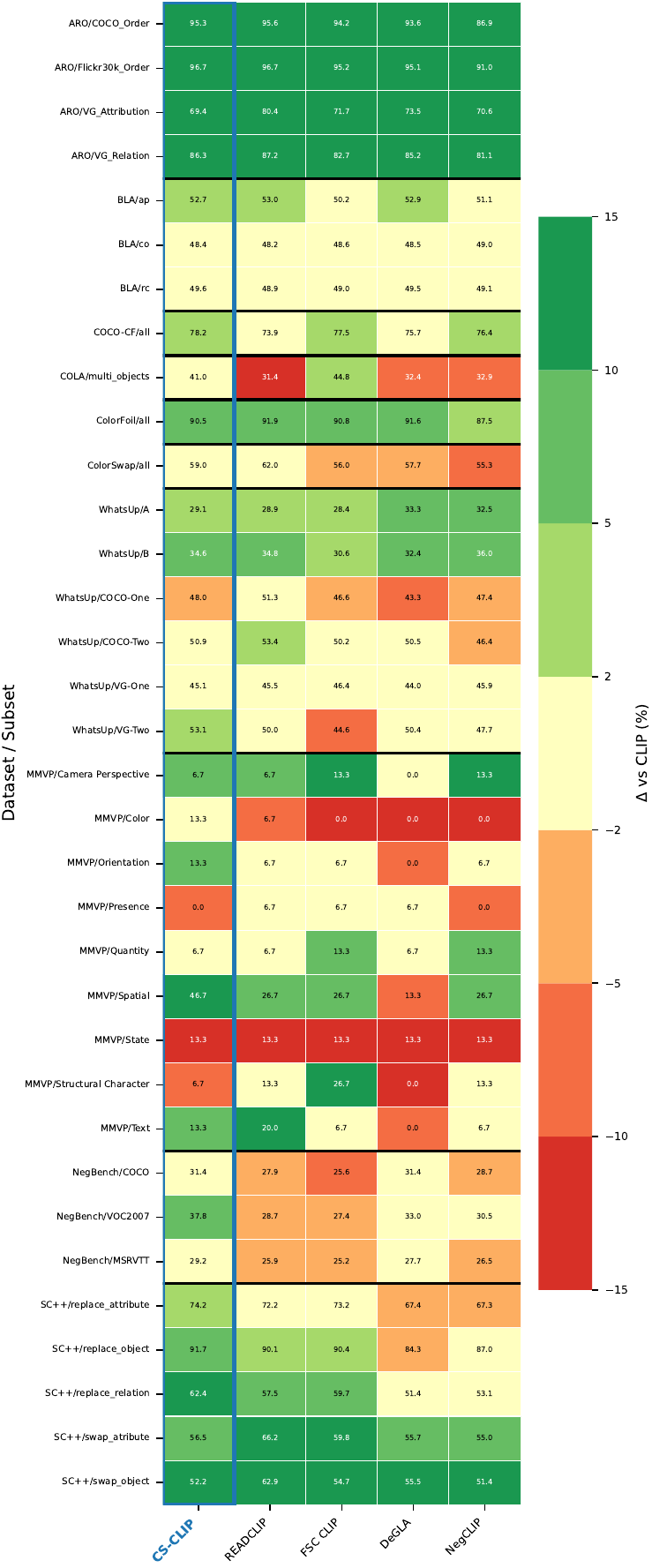}
  \caption{\textbf{Per-subset compositional I2T accuracy (Part 1/2).}
  Green: improvement over CLIP; red: degradation.}
  \label{fig:subset_heatmap_delta_part1}
\end{figure*}

\begin{figure*}[t]
  \centering
  \includegraphics[width=0.5\textwidth]{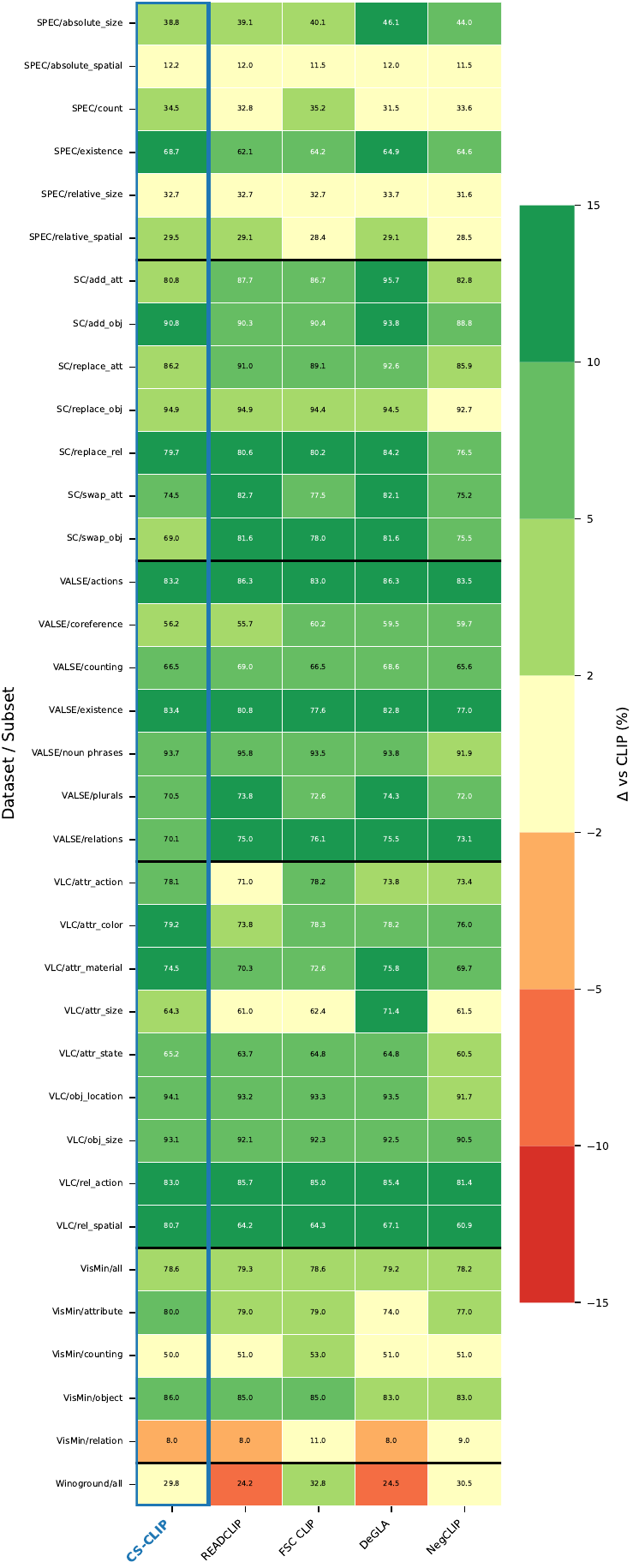}
  \caption{\textbf{Per-subset compositional I2T accuracy (Part 2/2).}
  Continuation of Figure~\ref{fig:subset_heatmap_delta_part1}.}
  \label{fig:subset_heatmap_delta_part2}
\end{figure*}

Figures~\ref{fig:subset_heatmap_delta_part1} and \ref{fig:subset_heatmap_delta_part2} 
visualize I2T accuracy differences relative to CLIP across all benchmark subsets, providing 
a comprehensive view of each model's strengths and weaknesses beyond aggregate scores.

\subsection{Per-Capability Results}
\label{app:compositional:per_capability}

\begin{table}[t]
  \centering
  \scriptsize
  \setlength{\tabcolsep}{2pt}
  \renewcommand{\arraystretch}{1.05}
  \caption{\textbf{Performance by sub-capability (I2T accuracy, \%).}
  Scores macro-averaged within each sub-capability.
  \textbf{Bold}: best; \underline{underline}: second best.}
  \label{tab:capability_sub}
  \resizebox{\textwidth}{!}{
  \begin{tabular}{p{4.2cm}ccccccccccc}
    \toprule
    Model & Attr. Bind. & Attr. Recog. & Coref. & Count & Exist. & Negation & Obj. Recog. & Pred. Sens. & Role Sens. & Syntax & Avg \\
    \midrule
    
    \multicolumn{12}{@{}l}{\textit{Pre-training variants (zero-shot)}} \\
    \midrule
    CLIP & 39.3 & 53.4 & 51.4 & 34.3 & 24.1 & 30.6 & 65.0 & 42.4 & 34.6 & 40.9 & 41.6 \\
    TripletCLIP (CC12M) & 35.5 & 48.4 & 50.9 & 26.9 & 24.1 & 27.3 & 56.6 & 42.7 & 33.9 & 35.3 & 38.2 \\
    LaCLIP (CC12M) & 37.7 & 47.5 & 52.2 & 29.6 & 29.0 & 25.5 & 55.3 & 41.1 & 35.3 & 38.1 & 39.1 \\
    \midrule
    
    \multicolumn{12}{@{}l}{\textit{Fine-tuned on COCO (direct comparison)}} \\
    \midrule
    \rowcolor{gray!10}\textbf{CS-CLIP} & 43.8\scalebox{0.6}{$\pm$1.0} & 58.1\scalebox{0.6}{$\pm$0.5} & 56.2\scalebox{0.6}{$\pm$1.7} & 37.7\scalebox{0.6}{$\pm$0.6} & \underline{30.7}\scalebox{0.6}{$\pm$1.2} & 32.8\scalebox{0.6}{$\pm$1.2} & \underline{71.3}\scalebox{0.6}{$\pm$0.7} & 51.0\scalebox{0.6}{$\pm$0.3} & \textbf{40.1}\scalebox{0.6}{$\pm$0.6} & \underline{54.3}\scalebox{0.6}{$\pm$0.6} & \underline{47.6}\scalebox{0.6}{$\pm$0.8} \\
    ReadCLIP & \textbf{46.7} & \underline{58.4} & 55.7 & \textbf{39.0} & \textbf{31.4} & 27.5 & \textbf{72.0} & \textbf{54.1} & \underline{38.8} & \textbf{55.5} & \textbf{47.9} \\
    FSC-CLIP (COCO) & \underline{45.3} & \textbf{58.8} & \underline{60.2} & \textbf{39.0} & 30.0 & 26.0 & 71.1 & \underline{52.8} & 37.4 & \underline{54.3} & 47.5 \\
    DeGLA & 40.7 & 56.8 & 59.5 & 35.9 & 27.7 & 30.7 & 68.4 & 50.1 & 35.5 & 51.4 & 45.7 \\
    NegCLIP & 42.4 & 56.0 & 59.7 & \underline{38.4} & 28.8 & 28.6 & 69.4 & 50.3 & 37.1 & 52.5 & 46.3 \\
    LabCLIP & 40.0 & 54.6 & 54.3 & 32.9 & 28.4 & 31.2 & 67.1 & 46.8 & 35.9 & 49.1 & 44.0 \\
    CE-CLIP & 37.8 & 56.2 & \textbf{60.6} & 37.0 & 30.1 & 35.7 & 66.2 & 46.4 & 34.8 & 46.0 & 45.1 \\
    DAC (LLM) & 38.6 & 52.9 & 54.6 & 31.3 & 26.5 & 35.9 & 65.5 & 44.8 & 36.2 & 45.9 & 43.2 \\
    DAC (SAM) & 39.3 & 52.4 & 53.3 & 31.6 & 28.7 & \textbf{37.2} & 66.0 & 45.1 & 35.2 & 45.8 & 43.5 \\
    \midrule
    
    \multicolumn{12}{@{}l}{\textit{Fine-tuned on larger/other datasets (CC3M, LAION, RedCaps)}} \\
    \midrule
    FSC-CLIP (CC3M) & 40.3 & 57.1 & 55.7 & 34.9 & 27.5 & 27.0 & 67.1 & 48.9 & 37.4 & 49.5 & 44.5 \\
    CLoVe & 39.8 & 52.3 & 49.1 & 34.6 & 30.0 & 25.2 & 64.7 & 51.1 & 37.0 & 50.7 & 43.5 \\
    FSC-CLIP (L+C) & 39.5 & 55.0 & 52.7 & 34.8 & 27.5 & 29.1 & 67.0 & 48.1 & 38.0 & 51.1 & 44.3 \\
    CLIC (RedCaps) & 42.6 & 56.6 & 47.7 & 34.4 & 25.7 & 32.4 & 69.0 & 50.1 & 35.9 & 48.4 & 44.3 \\
    CLIC (LAION) & 40.4 & 56.5 & 51.7 & 32.8 & 25.1 & 30.3 & 67.7 & 48.3 & 35.6 & 47.1 & 43.6 \\
    CON-CLIP & 38.1 & 52.2 & 45.8 & 33.7 & 26.7 & 34.8 & 66.9 & 42.4 & 33.0 & 38.1 & 41.2 \\
    TSVLC & 38.4 & 52.4 & 53.9 & 32.5 & 30.1 & \underline{36.5} & 65.1 & 45.2 & 34.5 & 44.6 & 43.3 \\
    \bottomrule
  \end{tabular}
  }
\end{table}

\paragraph{Analysis.}
Table~\ref{tab:capability_sub} shows capability-level performance patterns.
CS-CLIP achieves the best Role Sensitivity (40.1\%), substantially outperforming CLIP 
(34.6\%) and matching or exceeding all other COCO-trained methods.
This aligns with our approach: unit-level supervision explicitly targets relational 
structure by contrasting entities in different roles (e.g., "dog chasing cat" vs. 
"cat chasing dog").
CS-CLIP also ranks second on Object Recognition (71.3\%), Existence (30.7\%), and
Syntax (54.3\%), and is competitive on Attribute Binding (43.8\%), demonstrating
broad improvements.

The capability breakdown reveals systematic weaknesses across all models.
Counting remains challenging (24--39\%), likely because numerical information is poorly 
grounded in bag-of-visual-features representations.
Negation understanding is similarly weak (26--37\%), consistent with prior findings that 
CLIP-style models struggle with linguistic negation.
Object Recognition shows the highest absolute performance (55--72\%), suggesting entity-level 
grounding is more robust than relational or linguistic phenomena.

Comparing COCO-trained models to larger-dataset variants, we observe that COCO fine-tuning 
generally produces better compositional capabilities than training on CC3M or LAION.
For example, CS-CLIP (COCO) achieves 47.6\% average across capabilities, while FSC-CLIP 
(CC3M) achieves 44.5\% despite being trained on 30$\times$ more data.
This suggests compositional robustness depends more on alignment signal quality than raw 
dataset size.

\subsection{Bidirectional Performance on Paired Datasets}
\label{app:compositional:bidirectional}

\begin{table*}[t]
  \centering
  \scriptsize
  \setlength{\tabcolsep}{2.5pt}
  \renewcommand{\arraystretch}{1.05}
  \caption{\textbf{Bidirectional performance on paired compositional datasets (accuracy, \%).}
  I2T = Image-to-Text, T2I = Text-to-Image, Grp = Group accuracy (requires correctness in both directions).
  CSwap = ColorSwap, Wino.\ = Winoground.
  Checkpoints and within-dataset aggregation match the 16-benchmark summary;
  averages here give equal weight to the seven displayed datasets.
  \textbf{Bold}: best; \underline{underline}: second best at the displayed precision.}
  \label{tab:bidirectional}
  \resizebox{\textwidth}{!}{
  \begin{tabular}{l|ccc|ccc|ccc|ccc|ccc|ccc|ccc|ccc}
    \toprule
    & \multicolumn{3}{c|}{COCO-CF} & \multicolumn{3}{c|}{COLA} & \multicolumn{3}{c|}{CSwap} & \multicolumn{3}{c|}{MMVP} & \multicolumn{3}{c|}{SPEC} & \multicolumn{3}{c|}{VisMin} & \multicolumn{3}{c|}{Wino.} & \multicolumn{3}{c}{Average} \\
    Model & I2T & T2I & Grp & I2T & T2I & Grp & I2T & T2I & Grp & I2T & T2I & Grp & I2T & T2I & Grp & I2T & T2I & Grp & I2T & T2I & Grp & I2T & T2I & Grp \\
    \midrule
    
    \multicolumn{25}{@{}l}{\textit{Pre-training variants (zero-shot)}} \\
    \midrule
    CLIP & 74.0 & 73.1 & 60.4 & \underline{41.9} & 22.4 & 14.8 & \underline{60.7} & \textbf{72.7} & \textbf{45.0} & \underline{13.3} & \underline{9.6} & \textbf{5.2} & 32.7 & 31.8 & 14.2 & 57.9 & 50.1 & 38.0 & 29.8 & 10.8 & 7.8 & 44.3 & 38.6 & 26.5 \\
    TripletCLIP & 47.6 & 45.8 & 31.1 & 24.3 & \underline{26.2} & 11.4 & 55.7 & 64.0 & 38.0 & 10.4 & \underline{9.6} & \underline{3.7} & 30.5 & 28.8 & 13.1 & 35.8 & 32.6 & 15.2 & 22.2 & 6.2 & 3.8 & 32.3 & 30.5 & 16.6 \\
    LaCLIP & 41.0 & 36.7 & 23.9 & 21.9 & 14.8 & 6.7 & 56.7 & 63.7 & 40.3 & \textbf{15.6} & 4.4 & 3.0 & 29.2 & 28.2 & 11.5 & 33.7 & 27.7 & 13.0 & 26.2 & 8.0 & 6.2 & 32.0 & 26.2 & 14.9 \\
    \midrule
    
    \multicolumn{25}{@{}l}{\textit{Fine-tuned on COCO (direct comparison)}} \\
    \midrule
    \rowcolor{gray!10}\textbf{CS-CLIP (Ours)} & \underline{78.2} & \underline{75.9} & \textbf{65.1} & 41.0 & \underline{26.2} & \underline{20.0} & 59.0 & 69.3 & 40.7 & \underline{13.3} & \underline{9.6} & \underline{3.7} & \underline{36.1} & 31.9 & 15.0 & \underline{60.5} & 52.6 & 42.2 & 29.8 & \textbf{13.0} & 8.0 & \underline{45.4} & \underline{39.8} & \textbf{27.8} \\
    FSC-CLIP & 77.5 & 75.5 & \underline{64.5} & \textbf{44.8} & \textbf{27.1} & \textbf{21.9} & 56.0 & 69.0 & 38.3 & 12.6 & 7.4 & 0.7 & 35.3 & \underline{33.0} & 14.6 & \textbf{61.3} & \textbf{55.9} & \textbf{44.5} & \textbf{32.8} & 12.0 & \underline{9.5} & \textbf{45.8} & \textbf{40.0} & \underline{27.7} \\
    ReadCLIP & 73.9 & 75.6 & 62.7 & 31.4 & 25.7 & 15.2 & \textbf{62.0} & 67.3 & \underline{43.3} & 11.9 & 4.4 & 2.2 & 34.6 & 32.3 & 13.2 & \underline{60.5} & \underline{55.2} & \underline{44.2} & 24.2 & 10.2 & 6.0 & 42.6 & 38.7 & 26.7 \\
    DeGLA & 75.7 & 74.6 & 62.7 & 32.4 & 20.0 & 15.7 & 57.7 & 71.0 & 41.7 & 4.4 & 7.4 & 0.7 & \textbf{36.2} & 31.9 & \textbf{16.4} & 59.0 & 53.1 & 40.0 & 24.5 & 10.5 & 7.5 & 41.4 & 38.4 & 26.4 \\
    NegCLIP & 76.4 & 72.9 & 61.6 & 32.9 & 16.2 & 11.4 & 55.3 & 70.7 & 39.7 & 10.4 & \textbf{10.4} & 2.2 & 35.6 & \textbf{33.4} & 15.0 & 59.6 & 54.4 & 41.8 & 30.5 & 11.2 & 8.0 & 43.0 & 38.5 & 25.7 \\
    LabCLIP & 75.8 & 72.4 & 61.1 & 32.9 & \underline{26.2} & 16.7 & 57.7 & 62.7 & 37.0 & 11.1 & 8.1 & 2.2 & 33.5 & 30.6 & 12.9 & 54.8 & 47.6 & 35.2 & 24.5 & 10.2 & 6.8 & 41.5 & 36.8 & 24.6 \\
    CE-CLIP & 69.2 & 74.6 & 58.8 & 24.8 & 23.3 & 11.0 & 60.3 & 66.3 & 42.7 & 8.9 & 5.9 & 1.5 & 34.3 & 31.8 & 13.7 & 59.4 & 54.7 & 42.8 & 19.8 & 12.0 & 5.2 & 39.5 & 38.4 & 25.1 \\
    DAC (LLM) & 72.2 & 69.0 & 56.7 & 35.7 & 22.4 & 16.2 & 56.0 & 63.0 & 36.7 & 11.1 & \underline{9.6} & 2.2 & 33.0 & 29.8 & 14.1 & 51.5 & 45.9 & 32.2 & 25.5 & 9.8 & 6.5 & 40.7 & 35.6 & 23.5 \\
    DAC (SAM) & 72.8 & 68.4 & 56.5 & 34.3 & 20.5 & 14.8 & 54.7 & 65.7 & 36.0 & 8.9 & 8.1 & 0.7 & 32.2 & 29.8 & 13.6 & 51.2 & 45.4 & 32.2 & 26.5 & 10.2 & 6.5 & 40.1 & 35.4 & 22.9 \\
    \midrule
    
    \multicolumn{25}{@{}l}{\textit{Fine-tuned on larger/other datasets (CC3M, LAION, RedCaps)}} \\
    \midrule
    FSC-CLIP (CC3M) & 73.7 & 73.4 & 60.6 & 37.1 & 20.5 & 13.8 & 56.3 & 69.7 & 39.7 & 8.9 & 6.7 & 1.5 & 35.7 & 32.9 & \underline{15.2} & 59.0 & 54.6 & 42.0 & 29.2 & 8.8 & 6.0 & 42.9 & 38.1 & 25.5 \\
    FSC-CLIP (L+C) & 75.8 & \textbf{76.0} & 63.6 & 30.0 & 21.0 & 11.4 & 55.0 & \underline{72.3} & 39.0 & 9.6 & 4.4 & 1.5 & 34.3 & 32.5 & 14.3 & 57.8 & 52.7 & 40.0 & 27.8 & 9.2 & 6.8 & 41.5 & 38.3 & 25.2 \\
    CLoVe & 73.3 & 73.2 & 59.7 & 35.7 & 21.4 & 14.3 & 58.0 & 68.7 & 39.3 & \underline{13.3} & 3.7 & 0.7 & 32.0 & 31.5 & 13.5 & 53.5 & 48.1 & 34.2 & 22.5 & 10.2 & 6.2 & 41.2 & 36.7 & 24.0 \\
    CLIC (RedCaps) & 77.4 & 72.6 & 61.8 & 36.2 & 19.0 & 11.4 & 60.0 & 66.0 & 42.3 & 10.4 & \textbf{10.4} & 0.7 & 32.6 & 30.1 & 11.4 & 54.2 & 46.8 & 34.5 & \underline{32.2} & \underline{12.5} & \textbf{10.8} & 43.3 & 36.8 & 24.7 \\
    CLIC (LAION) & 78.0 & 72.5 & 62.4 & 34.3 & 24.3 & 14.3 & 58.3 & 66.7 & 40.7 & 10.4 & 5.2 & 0.7 & 33.0 & 30.3 & 11.7 & 55.4 & 47.6 & 35.2 & 31.5 & 11.8 & 9.2 & 43.0 & 36.9 & 24.9 \\
    CON-CLIP & \textbf{78.8} & 73.7 & 63.5 & 36.2 & 16.2 & 8.1 & 56.3 & 66.7 & 40.7 & 8.9 & \textbf{10.4} & 3.0 & 33.5 & 32.3 & 11.9 & 54.9 & 44.8 & 33.0 & 29.0 & 10.0 & 7.2 & 42.5 & 36.3 & 23.9 \\
    TSVLC & 73.2 & 68.9 & 56.9 & 33.3 & 17.6 & 11.0 & 54.3 & 65.7 & 37.7 & 11.9 & \underline{9.6} & 3.0 & 32.7 & 29.6 & 13.6 & 51.8 & 45.7 & 33.2 & 26.8 & 10.5 & 7.8 & 40.6 & 35.4 & 23.3 \\
    \bottomrule
  \end{tabular}
  }
\end{table*}

\paragraph{Analysis.}
Table~\ref{tab:bidirectional} reports bidirectional performance on seven paired compositional 
datasets that provide both image and text foils.
CS-CLIP achieves the \textbf{best average Group Accuracy} (27.8\%), indicating strong
performance when both retrieval directions must succeed simultaneously.
This confirms that unit-level supervision benefits both I2T and T2I matching rather than 
improving one direction at the expense of the other.

Among COCO-trained models, CS-CLIP and FSC-CLIP achieve comparable average I2T (45.4\% vs.\ 45.8\%)
and T2I (39.8\% vs.\ 40.0\%) accuracy, with Group Accuracy of 27.8\% and 27.7\%, respectively.
Notably, CS-CLIP achieves the best Winoground T2I score (13.0\%) among all models, 
demonstrating improved role sensitivity in the T2I direction.
Performance on MMVP remains challenging across all models (Group Accuracy below 6\%),
suggesting that fine-grained visual attribute distinctions require further advances 
beyond compositional training signals.

\subsection{Downstream Zero-Shot Classification and Retrieval}
\label{app:downstream}

We evaluate whether compositional improvements come at the cost of standard vision-language 
tasks.
Table~\ref{tab:downstream_zs_combined} reports zero-shot classification on ImageNet variants 
and fine-grained datasets, while Table~\ref{tab:downstream_retrieval} reports image--text 
retrieval on Flickr8k and MS-COCO.

\paragraph{Zero-shot classification.}
Given an image $I$ and class prompts $\{T_c\}$, we compute similarities $s(I,T_c)$ and 
predict the class with highest score.
We report Acc@1 and Acc@5 on seven datasets.

\paragraph{Image--text retrieval.}
Given an image (or text) query and a gallery of candidate texts (or images), we rank 
candidates by similarity and report Recall@1 in both directions: I2T (image-to-text) and 
T2I (text-to-image).

\begin{table*}[t]
\centering
\scriptsize
\caption{\textbf{Zero-shot classification (Acc@1 / Acc@5, \%).}
IN1k: ImageNet-1k \cite{deng2009imagenet}; INv2: ImageNetV2 \cite{recht2019imagenetv2}; Sketch: ImageNet-Sketch \cite{wang2019learning}; IN-O: ImageNet-O \cite{hendrycks2021natural}; 
Cal101: Caltech101 \cite{fei2004caltech101}; C10: CIFAR-10 \cite{krizhevsky2009tiny}; SUN: SUN397 \cite{xiao2010sun}.
\textbf{Bold}: best; \underline{underline}: second best.}
\label{tab:downstream_zs_combined}
\begin{adjustbox}{max width=\textwidth}
\setlength{\tabcolsep}{3.5pt}
\renewcommand{\arraystretch}{1.05}
\begin{tabular}{l|rr|rr|rr|rr|rr|rr|rr|rr}
\toprule
\multirow{2}{*}{\textbf{Model}} & \multicolumn{2}{c|}{\textbf{IN1k}} & \multicolumn{2}{c|}{\textbf{INv2}} & \multicolumn{2}{c|}{\textbf{Sketch}} & \multicolumn{2}{c|}{\textbf{IN-O}} & \multicolumn{2}{c|}{\textbf{Cal101}} & \multicolumn{2}{c|}{\textbf{C10}} & \multicolumn{2}{c|}{\textbf{SUN}} & \multicolumn{2}{c}{\textbf{Avg}} \\
\cmidrule(lr){2-3} \cmidrule(lr){4-5} \cmidrule(lr){6-7} \cmidrule(lr){8-9} \cmidrule(lr){10-11} \cmidrule(lr){12-13} \cmidrule(lr){14-15} \cmidrule(lr){16-17}
 & @1 & @5 & @1 & @5 & @1 & @5 & @1 & @5 & @1 & @5 & @1 & @5 & @1 & @5 & @1 & @5 \\
\midrule

\multicolumn{17}{@{}l}{\textit{Pre-training variants (zero-shot)}} \\
\midrule
CLIP & \underline{60.2} & 85.9 & \underline{52.4} & 79.4 & \textbf{46.4} & \textbf{72.6} & \textbf{50.7} & \textbf{82.3} & 81.9 & 94.1 & 88.4 & 99.7 & \textbf{65.0} & \underline{91.7} & \underline{63.6} & 86.5 \\
TripletCLIP (CC12M) & 23.0 & 46.9 & 19.3 & 42.2 & 14.8 & 33.3 & 29.2 & 57.8 & 65.5 & 89.1 & 63.2 & 95.5 & 36.4 & 68.5 & 35.9 & 61.9 \\
\midrule

\multicolumn{17}{@{}l}{\textit{Fine-tuned on COCO (direct comparison)}} \\
\midrule
\rowcolor{gray!10}\textbf{CS-CLIP} & 58.9 & 86.3 & 51.7 & 80.9 & 38.0 & 65.4 & 43.4 & 75.2 & 81.5 & \textbf{97.9} & 88.7 & 99.6 & 57.4 & 86.8 & 59.9 & 84.6 \\
FSC-CLIP (COCO) & 59.2 & 86.4 & 51.7 & 80.5 & 38.9 & 66.6 & 45.5 & 76.6 & 81.8 & 96.7 & 89.1 & 99.6 & 62.4 & 90.5 & 61.2 & 85.3 \\
NegCLIP & 55.8 & 83.8 & 48.9 & 78.0 & 35.9 & 62.6 & 42.4 & 72.7 & 81.2 & \underline{96.8} & 85.9 & 99.5 & 57.7 & 87.3 & 58.2 & 83.0 \\
DeGLA & 56.3 & 84.5 & 50.0 & 79.0 & 36.8 & 64.0 & 44.4 & 75.2 & 80.9 & 95.8 & 86.0 & 99.5 & 60.3 & 88.9 & 59.3 & 83.9 \\
ReadCLIP & 51.5 & 80.8 & 45.3 & 75.0 & 32.8 & 59.8 & 44.4 & 74.7 & 78.3 & 92.9 & 87.2 & 99.6 & 56.5 & 86.1 & 56.6 & 81.3 \\
CE-CLIP & 49.9 & 80.5 & 43.3 & 74.3 & 31.5 & 58.6 & 44.9 & 74.8 & 78.4 & 94.6 & 85.9 & 98.7 & 56.6 & 87.5 & 55.8 & 81.3 \\
DAC (LLM) & 52.7 & 81.3 & 46.3 & 75.2 & 35.5 & 62.9 & 43.7 & 71.9 & 79.3 & 93.5 & 87.0 & 98.8 & 56.3 & 87.2 & 57.3 & 81.5 \\
DAC (SAM) & 52.9 & 81.4 & 46.8 & 75.5 & 35.3 & 62.6 & 44.0 & 72.4 & 78.8 & 93.0 & 86.7 & 98.9 & 55.1 & 86.8 & 57.1 & 81.5 \\
\midrule

\multicolumn{17}{@{}l}{\textit{Fine-tuned on larger/other datasets (CC3M, LAION, RedCaps)}}\\
\midrule
CON-CLIP & \textbf{63.2} & \underline{88.7} & \textbf{55.5} & \underline{83.2} & \underline{42.8} & \underline{70.7} & 47.0 & 78.0 & \textbf{82.5} & 95.8 & \textbf{90.7} & \textbf{99.8} & 63.9 & \textbf{92.1} & \textbf{63.7} & \textbf{86.9} \\
CLIC (RedCaps) & 62.3 & 88.1 & 55.3 & 82.5 & 41.8 & 69.6 & 47.2 & 77.1 & \underline{82.4} & 96.7 & 89.8 & 99.6 & 64.2 & 91.6 & 63.3 & 86.5 \\
CLIC (LAION) & 61.7 & 87.7 & 55.0 & 82.4 & 41.4 & 69.1 & 46.8 & 77.7 & 81.8 & 96.1 & 89.5 & 99.6 & \underline{64.3} & 91.6 & 62.9 & 86.3 \\
FSC-CLIP (L+C) & 58.1 & 85.8 & 51.0 & 80.0 & 39.8 & 67.2 & \underline{48.3} & 76.8 & 79.9 & 93.7 & 87.5 & 99.2 & \underline{64.3} & 91.5 & 61.3 & 84.9 \\
FSC-CLIP (CC3M) & 54.9 & 83.8 & 47.7 & 78.4 & 37.6 & 65.0 & 46.0 & 74.9 & 80.3 & 94.2 & \underline{89.5} & 99.5 & 62.7 & 90.5 & 59.8 & 83.8 \\
TSVLC & 54.3 & 82.4 & 47.6 & 76.6 & 36.1 & 63.5 & 44.5 & 73.7 & 79.4 & 93.5 & 87.0 & 99.0 & 57.0 & 88.0 & 58.0 & 82.4 \\
CLoVe & 52.9 & 81.8 & 46.6 & 76.2 & 35.5 & 62.7 & 41.6 & 72.6 & 78.6 & 89.9 & 88.4 & \underline{99.7} & 60.2 & 88.7 & 57.7 & 81.6 \\
\bottomrule
\end{tabular}
\end{adjustbox}
\end{table*}

\begin{table}[t]
\centering
\scriptsize
\caption{\textbf{Image--text retrieval (Recall@1, \%).}
I2T: image-to-text; T2I: text-to-image.
\textbf{Bold}: best; \underline{underline}: second best.}
\label{tab:downstream_retrieval}
\begin{adjustbox}{max width=\columnwidth}
\setlength{\tabcolsep}{3pt}
\renewcommand{\arraystretch}{1.05}
\begin{tabular}{l|cc|cc|cc}
\toprule
& \multicolumn{2}{c|}{\textbf{Flickr8k}} & \multicolumn{2}{c|}{\textbf{MS-COCO}} & \multicolumn{2}{c}{\textbf{Average}} \\
\cmidrule(lr){2-3}\cmidrule(lr){4-5}\cmidrule(lr){6-7}
\textbf{Model} & I2T & T2I & I2T & T2I & I2T & T2I \\
\midrule

\multicolumn{7}{@{}l}{\textit{Pre-training variants}} \\
\midrule
CLIP & 56.2 & 72.9 & 34.3 & 52.5 & 45.2 & 62.7 \\
TripletCLIP (CC12M) & 23.4 & 28.4 & 11.3 & 14.9 & 17.3 & 21.7 \\
\midrule

\multicolumn{7}{@{}l}{\textit{Fine-tuned on COCO (direct comparison)}} \\
\midrule
\rowcolor{gray!10}\textbf{CS-CLIP} & \textbf{67.3} & \textbf{81.5} & \underline{46.3} & \textbf{61.9} & \textbf{56.8} & \textbf{71.7} \\
FSC-CLIP (COCO) & \underline{67.0} & \underline{79.2} & 46.0 & \underline{60.4} & \underline{56.5} & \underline{69.8} \\
ReadCLIP & 66.6 & 75.6 & \textbf{47.1} & 60.3 & \textbf{56.8} & 68.0 \\
DeGLA & 65.3 & 70.2 & 43.2 & 50.9 & 54.2 & 60.6 \\
NegCLIP & 63.9 & 75.6 & 41.5 & 56.2 & 52.7 & 65.9 \\
CE-CLIP & 64.8 & 67.8 & \textbf{47.1} & 56.1 & 55.9 & 61.9 \\
DAC (LLM) & 53.5 & 66.7 & 30.3 & 46.0 & 41.9 & 56.3 \\
DAC (SAM) & 52.0 & 66.5 & 29.8 & 45.9 & 40.9 & 56.2 \\
\midrule

\multicolumn{7}{@{}l}{\textit{Fine-tuned on larger/other datasets (CC3M, LAION, RedCaps)}} \\
\midrule
CLIC (RedCaps) & 56.1 & 70.8 & 34.5 & 47.1 & 45.3 & 58.9 \\
CLIC (LAION) & 56.5 & 68.3 & 34.6 & 47.6 & 45.5 & 57.9 \\
FSC-CLIP (CC3M) & 63.6 & 70.0 & 40.9 & 47.6 & 52.3 & 58.8 \\
FSC-CLIP (L+C) & 63.2 & 68.8 & 40.4 & 49.6 & 51.8 & 59.2 \\
CLoVe & 58.8 & 64.0 & 37.3 & 46.3 & 48.1 & 55.2 \\
CON-CLIP & 56.6 & 70.0 & 30.9 & 49.1 & 43.7 & 59.5 \\
TSVLC & 51.6 & 66.1 & 29.8 & 45.9 & 40.7 & 56.0 \\
\bottomrule
\end{tabular}
\end{adjustbox}
\end{table}

\paragraph{Analysis.}
CS-CLIP maintains competitive downstream performance while achieving the best compositional 
robustness.
On zero-shot classification, CS-CLIP achieves 59.9\% Acc@1 and 84.6\% Acc@5, compared to 
CLIP's 63.6\% and 86.5\%.
The 3.7-point Acc@1 gap represents a modest trade-off for substantial compositional gains
(+5.8 points on the compositional suite, +42.4 points on COCO (Qwen) Half-Truth Accuracy).
Notably, CS-CLIP achieves the best Caltech101 Acc@5 (97.9\%), suggesting unit-level 
supervision does not uniformly harm recognition.

For retrieval, CS-CLIP achieves the best performance among all evaluated models: 56.8\% I2T 
and 71.7\% T2I, outperforming CLIP by +11.6 and +9.0 points respectively.
This indicates that compositional improvements directly benefit retrieval tasks where 
fine-grained matching matters.
The gains are particularly pronounced on Flickr8k (67.3\% I2T, 81.5\% T2I), where captions 
are more descriptive and compositional structure is more salient.

Comparing COCO-trained models, we observe that methods achieving strong compositional 
robustness (CS-CLIP, FSC-CLIP, ReadCLIP) also excel at retrieval, while zero-shot 
classification performance varies.
This suggests retrieval and compositional evaluation measure related capabilities, both 
require fine-grained text-image alignment, while zero-shot classification depends more on 
coarse-grained category recognition.

\section{Additional Ablations}
\label{app:ablations_full}

This appendix reports extended ablations covering optimization hyperparameters and unit construction choices not included in the main paper due to space constraints.
The main ablations (Section~\ref{sec:exp:ablations}) compare backbone, loss weight, training signals, foil placement, and foil generators. Here we report construction and optimization ablations. Half-Truth metrics use COCO (Qwen).

All metrics are accuracies (\%) and higher is better.
We report three groups: \textbf{(i) Compositionality} (I2T/T2I/Group averaged across 16 benchmarks),
\textbf{(ii) Half-Truth Accuracy} (All/Entity/Relation), and
\textbf{(iii) Downstream} (zero-shot classification averaged across 7 datasets and retrieval on Flickr8k/COCO).

\begin{table*}[t]
\centering
\caption{\textbf{Unit construction hyperparameters.}
We vary the number of units per caption $N$, the fraction of relation-type units, and the fraction of swap-style foils.
\textbf{Gray}: baseline configuration.}
\label{tab:app_abls_construction}

\scriptsize
\setlength{\tabcolsep}{3.0pt}
\renewcommand{\arraystretch}{1.08}

\begin{tabular*}{\textwidth}{@{\extracolsep{\fill}}l|ccc|ccc|ccc@{}}
\toprule
\textbf{Setting} &
\multicolumn{3}{c|}{\textbf{Compositionality}} &
\multicolumn{3}{c|}{\textbf{Half-Truth}} &
\multicolumn{3}{c}{\textbf{Downstream}} \\
\cmidrule(r){2-4}\cmidrule(lr){5-7}\cmidrule(l){8-10}
& \textbf{I2T} & \textbf{T2I} & \textbf{Grp} &
\textbf{All} & \textbf{Ent} & \textbf{Rel} &
\textbf{ZS Avg} & \textbf{Retr T2I} & \textbf{Retr I2T} \\
\midrule

\multicolumn{10}{@{}l}{\textit{Number of units per caption $N$}} \\
$N{=}1$ & 57.4 & 39.1 & 27.4 & 76.6 & 74.4 & 78.8 & 59.5 & 70.7 & 56.3 \\
\rowcolor{gray!10}\textbf{$N{=}2$ (Ours)} & \textbf{57.8} & \textbf{39.8} & \textbf{27.8} & 76.4 & 73.6 & 79.2 & \textbf{59.9} & \textbf{71.7} & \textbf{56.8} \\
$N{=}3$ & 57.3 & 38.7 & 27.4 & 74.5 & 72.4 & 76.5 & 59.6 & 70.6 & 56.4 \\
$N{=}4$ & 57.5 & 39.0 & 27.2 & 77.4 & 73.2 & 81.6 & 58.9 & 70.2 & 56.5 \\
\midrule

\multicolumn{10}{@{}l}{\textit{Relation probability $p$ (fraction of relation-type units)}} \\
$p{=}0.0$ (entities only) & 56.7 & 39.7 & 27.5 & 58.2 & 75.6 & 40.8 & 59.1 & 70.6 & 56.5 \\
$p{=}0.25$ & 57.1 & 39.3 & 27.0 & 75.6 & 76.8 & 74.5 & 59.4 & 70.8 & 56.7 \\
$p{=}0.50$ & 57.1 & 39.2 & 26.4 & 77.4 & 78.0 & 76.9 & \textbf{60.4} & 71.5 & 56.7 \\
$p{=}0.75$ & 57.5 & 39.5 & 27.5 & 77.4 & 75.2 & 79.6 & 59.3 & 70.1 & 56.4 \\
\rowcolor{gray!10}\textbf{$p{=}1.0$ (Ours)} & \textbf{57.8} & \textbf{39.8} & \textbf{27.8} & 76.4 & 73.6 & 79.2 & 59.9 & \textbf{71.7} & \textbf{56.8} \\
\midrule

\multicolumn{10}{@{}l}{\textit{Swap probability (fraction of swap-style foils)}} \\
swap$=0.0$ (add/replace only) & 57.0 & 39.5 & 27.7 & 76.6 & 77.2 & 76.1 & 59.7 & 69.4 & 56.2 \\
swap$=0.5$ & 57.6 & 38.5 & 26.6 & 76.4 & 74.0 & 78.8 & 58.7 & 69.9 & 56.7 \\
\rowcolor{gray!10}\textbf{swap$=1.0$ (Ours)} & \textbf{57.8} & \textbf{39.8} & \textbf{27.8} & 76.4 & 73.6 & 79.2 & \textbf{59.9} & \textbf{71.7} & \textbf{56.8} \\
\bottomrule
\end{tabular*}
\end{table*}

\textbf{Unit construction choices.}
Using only entity units gives 75.6\% entity accuracy but 40.8\% relation accuracy.
Increasing the relation sampling probability improves relations, reaching 79.6\%
at $p=0.75$ and 79.2\% at $p=1.0$. Four units per caption reach 81.6\% relation
accuracy, while two give stronger compositional I2T and retrieval performance.
Swap probability also shifts this balance: relation accuracy rises from 76.1\%
without swaps to 79.2\% with full swap usage.

\begin{table*}[t]
\centering
\caption{\textbf{Optimization hyperparameters.}
We vary learning rate, weight decay, and batch size. \textbf{Gray}: baseline configuration.}
\label{tab:app_abls_optim}

\scriptsize
\setlength{\tabcolsep}{3.1pt}
\renewcommand{\arraystretch}{1.08}

\begin{tabular*}{\textwidth}{@{\extracolsep{\fill}}l|ccc|ccc|ccc@{}}
\toprule
\textbf{Setting} &
\multicolumn{3}{c|}{\textbf{Compositionality}} &
\multicolumn{3}{c|}{\textbf{Half-Truth}} &
\multicolumn{3}{c}{\textbf{Downstream}} \\
\cmidrule(r){2-4}\cmidrule(lr){5-7}\cmidrule(l){8-10}
& \textbf{I2T} & \textbf{T2I} & \textbf{Grp} &
\textbf{All} & \textbf{Ent} & \textbf{Rel} &
\textbf{ZS Avg} & \textbf{Retr T2I} & \textbf{Retr I2T} \\
\midrule

\multicolumn{10}{@{}l}{\textit{Learning rate}} \\
LR$=10^{-6}$ & 56.6 & 39.9 & 27.2 & 66.6 & 66.9 & 66.3 & \textbf{61.7} & 70.8 & 55.7 \\
LR$=2{\times}10^{-6}$ & 57.3 & 39.0 & 27.2 & 70.5 & 68.1 & 72.9 & 61.3 & 71.4 & 56.6 \\
\rowcolor{gray!10}\textbf{LR$=5{\times}10^{-6}$ (Ours)} & \textbf{57.8} & 39.8 & 27.8 & 76.4 & 73.6 & 79.2 & 59.9 & \textbf{71.7} & \textbf{56.8} \\
LR$=10^{-5}$ & 57.6 & 40.2 & \textbf{28.1} & 78.6 & 75.2 & 82.0 & 58.0 & 70.0 & 56.3 \\
LR$=2{\times}10^{-5}$ & 57.7 & \textbf{40.7} & \textbf{28.1} & 78.8 & 74.4 & 83.1 & 56.5 & 69.0 & 56.0 \\
\midrule

\multicolumn{10}{@{}l}{\textit{Weight decay}} \\
WD$=10^{-3}$ & 57.4 & 39.4 & 27.1 & 76.2 & 72.4 & 80.0 & 59.2 & 70.3 & 56.4 \\
WD$=5{\times}10^{-3}$ & 57.4 & 39.0 & 27.2 & 75.0 & 73.2 & 76.9 & 59.4 & 70.5 & 56.1 \\
\rowcolor{gray!10}\textbf{WD$=10^{-2}$ (Ours)} & \textbf{57.8} & \textbf{39.8} & 27.8 & 76.4 & 73.6 & 79.2 & \textbf{59.9} & \textbf{71.7} & \textbf{56.8} \\
WD$=2{\times}10^{-2}$ & 57.5 & 39.3 & 27.4 & 76.4 & 74.0 & 78.8 & 59.8 & 70.7 & 56.2 \\
WD$=5{\times}10^{-2}$ & 57.3 & 39.5 & 27.6 & 76.8 & 74.0 & 79.6 & 59.2 & 70.4 & 56.6 \\
\midrule

\multicolumn{10}{@{}l}{\textit{Batch size}} \\
BS$=32$ & 57.0 & 38.7 & 26.7 & 75.0 & 72.0 & 78.0 & \textbf{60.4} & 69.8 & 55.8 \\
BS$=64$ & 57.3 & 39.4 & 27.0 & 75.8 & 71.7 & 80.0 & 59.9 & 70.1 & 56.1 \\
\rowcolor{gray!10}\textbf{BS$=128$ (Ours)} & \textbf{57.8} & \textbf{39.8} & \textbf{27.8} & 76.4 & 73.6 & 79.2 & 59.9 & \textbf{71.7} & \textbf{56.8} \\
\bottomrule
\end{tabular*}
\end{table*}

\textbf{Optimization hyperparameters.}
Increasing the learning rate to $2\times10^{-5}$ reaches 78.8\% overall and
83.1\% relation Half-Truth accuracy, but lowers zero-shot classification to 56.5\%.
The default $5\times10^{-6}$ balances Half-Truth accuracy (76.4\%), zero-shot
classification (59.9\%), and T2I retrieval (71.7\%). Weight decay has smaller
effects. Batch sizes 32 and 64 reach 75.0\% and 75.8\% Half-Truth accuracy,
respectively, compared to 76.4\% with batch size 128.

\subsection{Control Configurations}
The foil-placement and Mistral-training controls use 25 epochs, seed 42, and
configured batch size 256; the main Qwen-trained model uses batch size 128.
The controls therefore compare practical training configurations rather than
isolating supervision placement or generator choice with every hyperparameter fixed.

\section{Assets and Licenses}
\label{app:licenses}

Table~\ref{tab:licenses} lists assets used for CS-CLIP training, evaluation construction,
and compositional evaluation, along with their licenses or service terms.
All assets are used strictly for non-commercial academic research in accordance with their respective terms.

\begin{table}[h]
  \centering
  \small
  \setlength{\tabcolsep}{4pt}
  \renewcommand{\arraystretch}{1.1}
  \caption{\textbf{Assets used in this work and their licenses.}}
  \label{tab:licenses}
  \resizebox{\textwidth}{!}{
  \begin{tabular}{llll}
    \toprule
    \textbf{Asset} & \textbf{Type} & \textbf{License} & \textbf{Usage} \\
    \midrule
    MS-COCO~\cite{lin2014coco}             & Dataset & CC BY 4.0          & Training and half-truth evaluation \\
    CC3M~\cite{sharma2018conceptual}       & Dataset & Freely usable with attribution (Google LLC) & OOD half-truth evaluation \\
    OpenAI CLIP ViT-B/32~\cite{radford2021learning} & Model weights & MIT License & Initialization for CS-CLIP \\
    OpenCLIP~\cite{cherti2023openclip}     & Codebase & MIT License        & Fine-tuning framework \\
    Qwen3-8B-AWQ~\cite{qwen3}              & LLM     & Apache 2.0         & Caption parsing pipeline (text only) \\
    \href{https://huggingface.co/Qwen/Qwen3-VL-30B-A3B-Instruct}{Qwen3-VL-30B-A3B-Instruct} & VLM & Apache 2.0 & Evaluation screening and training-foil audit \\
    Claude Opus 5 & Hosted model & Proprietary (Anthropic terms) & Evaluation review and replacement proposals \\
    \midrule
    \multicolumn{4}{@{}l}{\textit{Compositional evaluation benchmarks (all publicly available)}} \\
    ARO~\cite{yuksekgonul2022when}           & Benchmark & MIT License       & Evaluation only \\
    BLA~\cite{chen2023bla}                  & Benchmark & MIT License       & Evaluation only \\
    COCO-CF~\cite{le2023cococf}             & Benchmark & CC BY 4.0         & Evaluation only \\
    COLA~\cite{ray2024cola}                 & Benchmark & MIT License       & Evaluation only \\
    ColorFoil~\cite{samin2025colorfoil}     & Benchmark & MIT License       & Evaluation only \\
    ColorSwap~\cite{burapacheep2024colorswap} & Benchmark & MIT License     & Evaluation only \\
    MMVP~\cite{tong2024eyes}                & Benchmark & MIT License       & Evaluation only \\
    NegBench~\cite{alhamoud2025vision}      & Benchmark & MIT License       & Evaluation only \\
    SPEC~\cite{peng2024spec}                & Benchmark & Research-only (no explicit license file) & Evaluation only \\
    SugarCrepe~\cite{hsieh2023sugarcrepe}   & Benchmark & MIT License       & Evaluation only \\
    SugarCrepe++~\cite{dumpala2024sugarcrepepp} & Benchmark & MIT License + CC BY 4.0 & Evaluation only \\
    VALSE~\cite{parcalabescu2022valse}      & Benchmark & MIT License       & Evaluation only \\
    VisMin~\cite{awal2024vismin}            & Benchmark & CC BY 4.0         & Evaluation only \\
    VL-CheckList~\cite{zhao2022vlchecklist} & Benchmark & Research-only (no explicit license file) & Evaluation only \\
    What'sUp~\cite{kamath2023whats}         & Benchmark & MIT License       & Evaluation only \\
    Winoground~\cite{thrush2022winoground}  & Benchmark & Custom research-only license (gated) & Evaluation only \\
    \bottomrule
  \end{tabular}
  }
\end{table}

Claude is accessed as a hosted service under the applicable Anthropic
\href{https://www.anthropic.com/legal/consumer-terms}{Consumer Terms}
or \href{https://www.anthropic.com/legal/commercial-terms}{Commercial Terms}.
The authors manually verify evaluation candidates after the automated checks
(Appendix~\ref{app:half_truth_per_condition}).

\newpage
\section*{NeurIPS Paper Checklist}

\begin{enumerate}

\item {\bf Claims}
    \item[] Question: Do the main claims made in the abstract and introduction accurately reflect the paper's contributions and scope?
    \item[] Answer: \answerYes{}
    \item[] Justification: The abstract and introduction state all main claims (half-truth failure mode, unit-level supervision, HT-Acc 76.4\% on COCO (Qwen), best compositional I2T 57.8\%, best Group Accuracy) that are directly supported by the main-paper tables (half-truth and ablations) and the compositional results in Appendix~\ref{app:compositional}. Scope is bounded to COCO fine-tuning; OOD transfer is presented as supporting evidence, not a primary claim.

\item {\bf Limitations}
    \item[] Question: Does the paper discuss the limitations of the work performed by the authors?
    \item[] Answer: \answerYes{}
    \item[] Justification: Section~\ref{sec:conclusion} discusses text-only parsing errors, the zero-shot classification trade-off, and limits on factual correctness and demographic fairness. Appendix~\ref{app:half_truth_per_condition} reports the training-foil audit and distinguishes automated training-data filtering from manual evaluation verification.

\item {\bf Theory assumptions and proofs}
    \item[] Question: For each theoretical result, does the paper provide the full set of assumptions and a complete (and correct) proof?
    \item[] Answer: \answerNA{}
    \item[] Justification: The paper makes no theoretical claims and contains no theorems, lemmas, or proofs; all contributions are empirical.

    \item {\bf Experimental result reproducibility}
    \item[] Question: Does the paper fully disclose all the information needed to reproduce the main experimental results of the paper to the extent that it affects the main claims and/or conclusions of the paper (regardless of whether the code and data are provided or not)?
    \item[] Answer: \answerYes{}
    \item[] Justification: Section~5 details the training setup (optimizer, learning rate, batch size, backbone, temperature). The caption-parsing and foil-generation pipeline is described in Section~3 and Appendix~\ref{app:llm_parsing}. All evaluation benchmarks are publicly available. Code is available at \url{https://github.com/kargibora/CS-CLIP}.

\item {\bf Open access to data and code}
    \item[] Question: Does the paper provide open access to the data and code, with sufficient instructions to faithfully reproduce the main experimental results, as described in supplemental material?
    \item[] Answer: \answerYes{}
    \item[] Justification: Code, environment requirements, and training and evaluation commands are available at \url{https://github.com/kargibora/CS-CLIP}. The repository links to the released checkpoint, manually verified Half-Truth evaluation set, and generated training annotations on Hugging Face; training images are obtained separately from COCO.

\item {\bf Experimental setting/details}
    \item[] Question: Does the paper specify all the training and test details (e.g., data splits, hyperparameters, how they were chosen, type of optimizer) necessary to understand the results?
    \item[] Answer: \answerYes{}
    \item[] Justification: Section~5 specifies the optimizer, learning rate, weight decay, batch size, training epochs, backbone, and temperature initialization; Section~\ref{sec:exp:ablations} examines the hyperparameter choices. Appendix~\ref{app:half_truth} describes the constructed evaluation sets, and Appendix~\ref{app:compositional:metrics} defines the compositional evaluation metrics.

\item {\bf Experiment statistical significance}
    \item[] Question: Does the paper report error bars suitably and correctly defined or other appropriate information about the statistical significance of the experiments?
    \item[] Answer: \answerYes{}
    \item[] Justification: Appendix~\ref{app:compositional:per_benchmark} reports per-benchmark standard deviations across five CS-CLIP runs. Point estimates use the best-performing run, with overall compositional I2T accuracy of 57.8\%. Baselines use their published checkpoints.

\item {\bf Experiments compute resources}
    \item[] Question: For each experiment, does the paper provide sufficient information on the computer resources (type of compute workers, memory, time of execution) needed to reproduce the experiments?
    \item[] Answer: \answerNo{}
    \item[] Justification: Section~5 reports training on 8$\times$A100 GPUs for 25 epochs with batch size 128. We do not report per-experiment wall-clock times, memory usage, or the total compute used across the research project.
    
\item {\bf Code of ethics}
    \item[] Question: Does the research conducted in the paper conform, in every respect, with the NeurIPS Code of Ethics \url{https://neurips.cc/public/EthicsGuidelines}?
    \item[] Answer: \answerYes{}
    \item[] Justification: We have reviewed the NeurIPS Code of Ethics. We fine-tune a publicly available vision-language model on COCO and evaluate on existing benchmarks and manually verified COCO/CC3M comparisons; we collect no private data and conduct no research with human participants.

\item {\bf Broader impacts}
    \item[] Question: Does the paper discuss both potential positive societal impacts and negative societal impacts of the work performed?
    \item[] Answer: \answerYes{}
    \item[] Justification: Section~\ref{sec:conclusion} discusses improved sensitivity to incorrect details in image--text retrieval, along with remaining factual errors and demographic biases. These limitations remain relevant when using or deploying the released model.
    
\item {\bf Safeguards}
    \item[] Question: Does the paper describe safeguards that have been put in place for responsible release of data or models that have a high risk for misuse (e.g., pre-trained language models, image generators, or scraped datasets)?
    \item[] Answer: \answerNA{}
    \item[] Justification: We release a fine-tuned CLIP checkpoint and annotations derived from existing image--caption datasets for retrieval research. The model card and dataset documentation describe their intended use and limitations; CS-CLIP does not guarantee factual correctness or demographic fairness.

\item {\bf Licenses for existing assets}
    \item[] Question: Are the creators or original owners of assets (e.g., code, data, models), used in the paper, properly credited and are the license and terms of use explicitly mentioned and properly respected?
    \item[] Answer: \answerYes{}
    \item[] Justification: Appendix~\ref{app:licenses} lists licenses and usage for the base datasets, CLIP weights, OpenCLIP code, and compositional benchmarks. The original sources are cited in the paper, and the released assets include their own documentation.

\item {\bf New assets}
    \item[] Question: Are new assets introduced in the paper well documented and is the documentation provided alongside the assets?
    \item[] Answer: \answerYes{}
    \item[] Justification: The Half-Truth benchmark contains 1,437 manually verified comparisons from COCO and CC3M, documented at \url{https://huggingface.co/datasets/kbora/Half-Truths}. The checkpoint and automatically generated training annotations are documented at \url{https://huggingface.co/kbora/CS-CLIP} and \url{https://huggingface.co/datasets/kbora/CS-CLIP-Training}, respectively; Appendix~\ref{app:half_truth_per_condition} distinguishes evaluation verification from training-data filtering.

\item {\bf Crowdsourcing and research with human subjects}
    \item[] Question: For crowdsourcing experiments and research with human subjects, does the paper include the full text of instructions given to participants and screenshots, if applicable, as well as details about compensation (if any)? 
    \item[] Answer: \answerNA{}
    \item[] Justification: We do not conduct any crowdsourcing or research with human subjects. The half-truth evaluation suite (1,437 comparisons across COCO and CC3M) is manually verified by the authors themselves; no crowdworkers or external annotators are involved.

\item {\bf Institutional review board (IRB) approvals or equivalent for research with human subjects}
    \item[] Question: Does the paper describe potential risks incurred by study participants, whether such risks were disclosed to the subjects, and whether Institutional Review Board (IRB) approvals (or an equivalent approval/review based on the requirements of your country or institution) were obtained?
    \item[] Answer: \answerNA{}
    \item[] Justification: No human subjects are involved in this research. The authors manually verified the evaluation foils against their images after automated screening. Training foils were automatically generated and filtered, without manual verification at scale.

\item {\bf Declaration of LLM usage}
    \item[] Question: Does the paper describe the usage of LLMs if it is an important, original, or non-standard component of the core methods in this research? Note that if the LLM is used only for writing, editing, or formatting purposes and does \emph{not} impact the core methodology, scientific rigor, or originality of the research, declaration is not required.
    \item[] Answer: \answerYes{}
    \item[] Justification: We use text-only Qwen3-8B-AWQ to parse captions and generate foils, with Mistral-8B in the cross-generator controls (Appendix~\ref{app:llm_parsing}). Evaluation candidates undergo image-grounded Qwen3-VL-30B screening and Claude Opus 5 review with proposed replacements, followed by manual verification by the authors (Appendix~\ref{app:half_truth_per_condition}); training foils are not manually verified at scale.

\end{enumerate}

\end{document}